\documentclass[journal]{IEEEtai}

\usepackage[colorlinks,urlcolor=blue,linkcolor=blue,citecolor=blue]{hyperref}

\usepackage{color,array}

\let\labelindent\relax
\usepackage{enumitem}
\usepackage{graphicx}
\usepackage{amsmath,amssymb,amsfonts}
\usepackage{booktabs}       
\usepackage{algorithmic}
\usepackage{graphicx}
\usepackage{textcomp}
\usepackage{enumitem}
\usepackage{caption}
\usepackage{subcaption}
\usepackage{makecell}
\usepackage{float}
\usepackage{xcolor}
\usepackage{svg}

\begin{document}

\title{SANSformers: Self-Supervised Forecasting in {\sc Electronic Health Records} with Attention-Free Models}

\author{Yogesh Kumar, Alexander Ilin, Henri Salo, Sangita Kulathinal, Maarit K. Leinonen, and Pekka Marttinen
\thanks{PM has received funding from the Research Council of Finland (grants no. 286607, 319323, 336033, 315896) BusinessFinland (884/31/2018), EU H2020 (101016775), and the Research Council of Finland Flagship program: Finnish Center for Artificial Intelligence (FCAI). We acknowledge the computational resources provided by Aalto Science-IT project.}
\thanks{Y. Kumar, A. Ilin and P. Marttinen are with the Department of Computer Science, Aalto University, Espoo, Finland (e-mail: yogesh.kumar@aalto.fi). }
\thanks{H. Salo and M. K. Leinonen are with the Information Services Department, Finnish Institute for Health and Welfare, Finland}
\thanks{S. Kulathinal is with the Department of Mathematics and Statistics, University of Helsinki, Finland}}


\maketitle

\begin{abstract}
    Despite the proven effectiveness of Transformer neural networks across multiple domains, their performance with Electronic Health Records (EHR) can be nuanced. The unique, multidimensional sequential nature of EHR data can sometimes make even simple linear models with carefully engineered features more competitive. Thus, the advantages of Transformers, such as efficient transfer learning and improved scalability are not always fully exploited in EHR applications. Addressing these challenges, we introduce SANSformer, an attention-free sequential model designed with specific inductive biases to cater for the unique characteristics of EHR data.

    In this work, we aim to forecast the demand for healthcare services, by predicting the number of patient visits to healthcare facilities. The challenge amplifies when dealing with divergent patient subgroups, like those with rare diseases, which are characterized by unique health trajectories and are typically smaller in size. To address this, we employ a self-supervised pretraining strategy, Generative Summary Pretraining (GSP), which predicts future summary statistics based on past health records of a patient. Our models are pretrained on a health registry of nearly one million patients, then fine-tuned for specific subgroup prediction tasks, showcasing the potential to handle the multifaceted nature of EHR data.
    
    In evaluation, SANSformer consistently surpasses robust EHR baselines, with our GSP pretraining method notably amplifying model performance, particularly within smaller patient subgroups. Our results illuminate the promising potential of tailored attention-free models and self-supervised pretraining in refining healthcare utilization predictions across various patient demographics.
\end{abstract}

\begin{IEEEImpStatement}
Large neural networks have demonstrated success in various predictive tasks using Electronic Health Records (EHR). However, their performance in small divergent patient cohorts, such as those with rare diseases, often falls short of simpler linear models due to the substantial data requirements of large models. To address this limitation, we introduce the SANSformers architecture, specifically designed for forecasting healthcare utilization within EHR. Distinct from traditional transformers, SANSformers utilize attention-free mechanisms, thereby reducing complexity. We also present Generative Summary Pretraining (GSP), a self-supervised learning technique that enables large neural networks to maintain predictive efficiency even with smaller patient subgroups. Through extensive evaluation across two real-world datasets, we provide a comparative analysis with existing state-of-the-art EHR prediction models, offering a new perspective on predicting healthcare utilization.
\end{IEEEImpStatement}


\section{Introduction}
\label{sec:intro}

\IEEEPARstart{T}{he} remarkable success of Transformer-based architectures \cite{vaswani2017attention} in Natural Language Processing (NLP) and Computer Vision benchmarks \cite{wang2018glue, deng2009imagenet} has extended to certain applications within the realm of Electronic Health Records (EHR). In instances where EHRs encompass natural language input, such as discharge notes written by healthcare professionals, Transformers have achieved impressive results \cite{yan2022clinical, jiang2023health}. Nonetheless, the unique structure and characteristics of EHR data, especially sequences of clinical codes, present challenges that can sometimes hinder Transformers from consistently outperforming simpler models with carefully engineered features \cite{ahmad2018death, bellamy2020evaluating, kelly2019key}. While Transformers excel in many areas, their performance with EHR can be nuanced, often requiring extensive pretraining on large datasets. This can render them less practical for applications with smaller or specialized datasets.

Clinical code sequences in EHRs exhibit unique properties in contrast to typical text or image data \cite{rasmy2021med}. For example, while consecutive words in a sentence or pixels in an image are usually strongly correlated, the consecutive visits in an EHR dataset can be disparate in terms of time and context. These visits may relate to different health issues and can be spaced years apart. 
This characteristic discrepancy has resulted in situations, where simpler models such as linear regression and recurrent neural networks (RNNs) still remain competitive\cite{razavian2015population,choi2016retain,kodialam2021deep}. Motivated by this observation, we hypothesize that the computationally and memory intensive self-attention mechanism in Transformers might be too complex for this application, leading us to explore a simpler yet effective alternative. In this paper, we propose a novel sequential architecture for EHR analysis inspired by the recent works that eliminate the need for recurrent, convolution or self-attention mechanisms\cite{liu2021pay,tolstikhin2021mlp}.

Our model, which we name SANSformers (for \textbf{S}equential \textbf{A}nalysis with \textbf{N}on-attentive \textbf{S}tructures for \textbf{e}lectronic health \textbf{r}ecords), is specifically designed with inductive biases that accommodate the unique features of EHR data. In particular, we incorporate axial mixing to address the inherent multi-dimensionality of clinical codes within each visit, and $\Delta\tau$ embeddings to accurately capture the time lapses between visits. This work thus marks a critical departure from conventional Transformer applications and offers a novel and potentially more efficient way to analyze EHR data.

\begin{figure}[]
  \centering
    \includegraphics[width=0.8\linewidth]{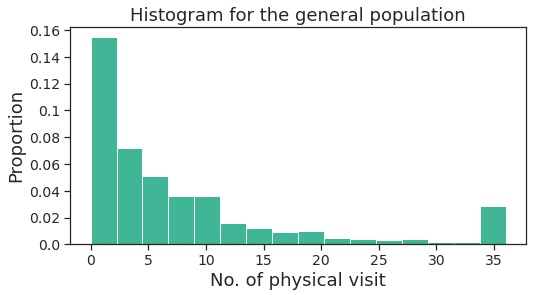}
    \label{fig:div-subgroups-sub1}
    
    \bigbreak

    \includegraphics[width=0.8\linewidth]{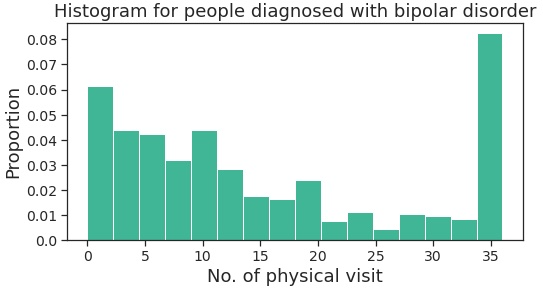}
    \label{fig:div-subgroups-sub2}
    
      \caption{\textbf{Highlighting the challenge of predicting with a divergent subgroup.}  The histograms depict the number of hospital visits for two groups: the general population (\emph{Top}) and individuals diagnosed with bipolar disorder (\emph{Bottom}). The observed bimodality in the histogram of counts results from the application of a \textit{topcap} measure, a mechanism designed to address the long-tail problem by capping the maximum count value at 36 visits. This highlights the disparate visitation patterns within subgroups, demonstrating the complexity in predicting future healthcare utilization. A Student's t-test indicates a significant difference between the distributions of general population and bipolar disorder (p-value $< 0.001$).}
  \label{fig:div-subgroups}
\end{figure}

Our overarching motivation is to construct a model capable of predicting future healthcare utilization -- a measure of an individual's use of medical services, such as hospital stays, doctor visits, and medical procedures -- based on their disease history and other pertinent factors. Such prediction models are key to efficient management and planning in healthcare \cite{tawhid2021machine,mizan2022medical}, and are for example already employed in numerous countries to allocate healthcare resources \cite{mcguire2018regulated}.
One significant challenge lies in making accurate predictions for patients belonging to divergent subgroups. These subgroups are characterized by differences in the distribution of the dependent variable. For instance, patients diagnosed with a specific disease often exhibit health history trajectories that deviate from those of other patients, as well as their own past patterns. This divergence can be particularly pronounced in cases of severe or chronic illnesses. Existing health economics research has addressed this challenge by developing models for specific subgroups \cite{shrestha2018mental,ellis2018risk}. However, building specialized models for each subgroup is not feasible when the subgroup size is small \cite{haendel2020many}, and training a single model for the entire population could yield inaccurate results for divergent subgroups.

Addressing this issue, we borrow a successful strategy from NLP and computer vision, where models pre-trained on larger corpora show improved data efficiency when fine-tuned for specific tasks \cite{pmlr-v27-bengio12a, devlin2019bert, radford2019language, dosovitskiy2020image, chen2020generative}. EHR datasets are known to be noisy \cite{russell2021electronic, miotto2016deep}, posing a potential issue for models such as GPT \cite{radford2019language} that are trained to predict the next token. In the presence of noise, these models may end up fitting on the noise instead of the actual structure of the data. To mitigate this issue, we apply the principle of pre-training in the EHR context, introducing a self-supervised regime—Generative Summary Pre-training (GSP)—that predicts summary statistics for a future window in the patient's history, such as the number of visits in the next year, based on current and past inputs. By predicting summary statistics, the effects of noise are alleviated, allowing for more reliable pre-training.

In our evaluation, we focus on patients diagnosed with Type 2 diabetes, bipolar disorder and multiple sclerosis, whose visitation patterns diverge strongly from the broader population ($p< 0.001$, t-test), as illustrated in Fig. \ref{fig:div-subgroups}, which features the bipolar disorder subgroup. This underscores the complexity and variability of health trajectories across different subgroups, highlighting the need for flexible and adaptable predictive models, such as our SANSformers.

Our main contributions are summarized as follows:

\begin{itemize}
\item We introduce SANSformers, a novel, attention-free sequential model specifically designed for EHR data, supplemented with inductive biases such as axial decomposition and $\Delta \tau$ embeddings to cater to the unique challenges posed by the EHR domain.
\item We conduct extensive comparisons of SANSformers with strong baseline models on two real-world EHR datasets. Our results highlight the superior data efficiency and prediction accuracy of our model compared to the existing baselines.
\item We demonstrate the value of self-supervised pre-training on a larger population for predicting future healthcare utilization of smaller, distinct patient subgroups. We introduce Generative Summary Pre-training (GSP), a self-supervised pre-training objective for EHR data, predicting future summary statistics. This offers considerable potential in improving healthcare resource allocation predictions, an application of significant importance in healthcare management.
\end{itemize}

\section{Related Work}

\textbf{Deep Learning on EHR.} Deep learning has been a focal point in EHR research, with diverse approaches being proposed. For instance, Lipton et al. \cite{lipton2015learning} used Long Short-Term Memory (LSTM) networks \cite{hochreiter1997long} for EHR phenotyping, applying sequential real-valued measurements of 13 vital signs to predict one of 128 diagnoses. Choi et al. \cite{choi2016doctor} proposed a multi-task learning framework using Gated Recurrent Units (GRU) \cite{cho2014learning} for phenotyping, showcasing the benefits of knowledge transfer from larger to smaller datasets. They also developed a bidirectional attention-based model, RETAIN, to enhance model interpretability \cite{choi2016retain}. Harutyunyan et al. \cite{harutyunyan2019multitask} contributed a benchmark framework for evaluating EHR models.

Notably, simpler models, such as linear ones, often exhibit competitive performance on EHR data \cite{razavian2015population,ahmad2018death,avati2018improving,miotto2016deep}. These models usually rely on manual feature engineering to construct patient state vectors from longitudinal health data, as opposed to training end-to-end from raw EHR data. In contrast, SANSformers seeks to directly leverage raw longitudinal EHR data, aiming to circumvent the need for extensive feature engineering while maintaining the simplicity and efficacy of non-attention-based architectures.

\begin{figure}[]
    \centering
    \includegraphics[scale=0.4]{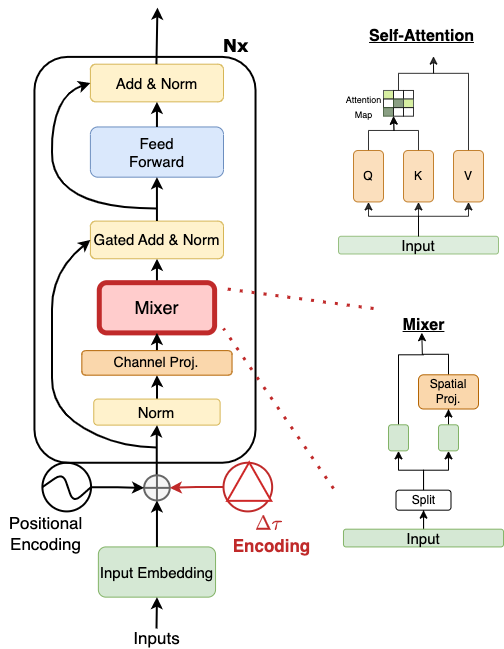}
    \caption{\textbf{SANSformer Architecture}. The figure provides a schematic representation of a single SANSformer layer. To emphasize the specific modifications introduced in our approach, we mirror the schematic of the Transformer architecture from Vaswani et al. \cite{vaswani2017attention}. Alterations from the conventional Transformer layer are highlighted in red, particularly the introduction of \(\Delta \tau\) embeddings and the replacement of the self-attention mechanism with our attention-free mixers. A side-by-side comparison on the right shows the conventional self-attention and Mixer, underscoring the efficiency achieved with fewer projection operations. Unlike the self-attention's three projections (to query, key, and value), the mixer employs just a channel and a spatial projection. For an exhaustive schematic of the entire architecture, please refer to Fig. \ref{fig:detailed-mixer}.
                }
    \label{fig:sansformer-fig}
\end{figure}

\textbf{Attention-less MLP Models.} A recent wave of research has questioned the necessity of attention mechanisms in Transformers, especially within the domain of computer vision. For instance, the MLP-Mixer proposed by Tolstikhin et al. \cite{tolstikhin2021mlp} replaced attention with ``mixing'' operations, and Liu et al. \cite{liu2021pay} introduced spatial gating mechanisms to further simplify the architecture. Other studies echo these findings \cite{touvron2021resmlp,lee2021fnet,melas2021you}. These developments challenge the indispensability of expensive self-attention mechanisms and motivate the extension of such models to EHR data. SANSformers represents, to our knowledge, the first endeavor to apply attention-less Transformers to EHR data. Fig. \ref{fig:sansformer-fig} highlights the distinctions of SANSformers from regular Transformers. While we replaced the self-attention mechanism, retaining other proven components from the Transformer model was a mindful decision to preserve stability and performance in EHR applications.

\textbf{Self-supervised pre-training.} Pioneering work by Howard et al. \cite{howard2018universal} and Peters et al. \cite{peters2018deep} demonstrated the potential of unsupervised language model pre-training to enhance accuracy in downstream tasks. However, the most remarkable improvements were observed when pre-training the Transformer architectures \cite{devlin2019bert, radford2019language, brown2020language, raffel2019exploring, liu2019roberta}.

Applying pre-trained Transformers to EHR data has shown promise, as evidenced by studies from Li et al. \cite{li2020behrt} and Rasmy et al. \cite{rasmy2021med}, who achieved improved performance over recurrent neural networks (RNNs) using pre-trained BERT Transformers. Their work only considered single discrete sequences, specifically diagnosis codes. Meng et al. \cite{meng2021bidirectional} further extended this to multi-modal data by using a topic model to also utilize the patient history from clinical notes. While the direct application of the BERT-based Masked Language Model (MLM) pre-training to clinical codes data has yielded encouraging results, this strategy lacks customization for EHR data or specific tasks at hand. Consequently, we propose our Generative Summary Pre-training (GSP) strategy, which is purposefully designed to better leverage the unique structure and characteristics of EHR data. Our proposed SANSformers model broadens this approach by incorporating diagnoses, procedures, and patient demographics. 

Shang et al. \cite{shang2019pre} introduced a unique approach, G-BERT, for medication recommendation. Their model combined Graph Neural Networks (GNNs) with BERT to efficiently represent medical codes, accounting for hierarchical structures. G-BERT was pre-trained on single-visit patient data, then fine-tuned on longitudinal data, achieving superior performance on the medication recommendation task. This model aligns with SANSformer's self-supervised pre-training approach, although SANSformer focuses on attention-less architectures.

Moreover, the strategy of reverse distillation, which initializes deep models using high-performing linear models, has shown notable success in clinical prediction tasks. The Self Attention with Reverse Distillation (SARD) architecture \cite{kodialam2021deep}, designed explicitly for insurance claims data, employs a mix of contextual and temporal embedding along with self-attention mechanisms, achieving superior performance on a variety of clinical prediction tasks, largely attributable to reverse distillation.

An extensive review by Krishnan et al. \cite{krishnan2022self} underscored the immense potential of self-supervised learning methods in healthcare. These methods can effectively leverage large-scale unannotated data across various modalities such as electronic health records, medical images, bioelectrical signals, and gene and protein sequences. This review highlights the value of these methods in handling multimodal datasets and addresses the challenges related to data bias, reinforcing the suitability of self-supervised learning for EHR data and its promising role in the advancement of medical AI.

\textbf{Handling Intra-visit Data in EHR.} Handling intra-visit data, the clinical codes arising from a single patient visit, presents a critical challenge in deep learning on EHR data. Common methods include summing or flattening codes across the intra-visit axis, but this may lead to a loss of nuanced information \cite{li2020behrt}.
Choi et al. \cite{choi2019graph} proposed a convolutional transformer model that incorporates an inductive bias for modeling interactions between different codes within a single visit. This concept was extended by Kumar et al. \cite{kumar2020predicting}, who used a $1 \times 1$ convolution to model these interactions, preserving more details from the intra-visit data.
Our approach, SANSformers, tackles this issue by leveraging axial decomposition, allowing us to model interactions within intra-visit data without resorting to flattening or summing. This approach retains critical information that could enhance prediction accuracy.

\section{Experimental setup}
\subsection{Patient cohorts}
We used two Electronic Health Records (EHRs) data sources for our experiments: a confidential dataset (Pummel) that was sourced from the Care Register for Health Care and Register of Primary Health Care visits maintained by the Finnish Institute for Health and Welfare (THL) and a smaller publicly available MIMIC-IV dataset, on which the readers can run our method and replicate our findings. \footnote{Code for SANSformer architectures can be found at: \url{https://github.com/ykumards/sansformers}} Table \ref{tab:basic-ehr-stats} lists some basic statistics of both datasets.

\begin{table}[]
  \caption{\textbf{Basic statistics of the two EMR datasets.}}
  \label{tab:basic-ehr-stats}
  \centering
  \small
  \begin{tabular}{@{}lrr@{}}
  \toprule
                                & \multicolumn{1}{c}{\textbf{Pummel}} & \multicolumn{1}{l}{\textbf{MIMIC-IV}} \\ \midrule
  No. of unique patients         & 1,050,512                         & 256,878                               \\
  No. of visits                  & 60,896,305                        & 523,740                               \\
  Avg. no. of visits per patient & 57.94                             & 2.04                                  \\
  Max no. of visits              & 1,443                             & 238                                   \\
  Avg no. of codes per visit     & 8.46                              & 18.84                                 \\
  Max no. of codes per visit     & 164                               & 110                                   \\
  Token Vocabulary Size          & 5237                              & 5019 \\ \bottomrule

  \end{tabular}
\end{table}

\subsubsection{Pummel}

The Pummel dataset encompasses the pseudonymized EHRs of all Finnish citizens aged 65 or older who have engaged with either primary or secondary healthcare services. This data spans seven years (2012 to 2018) and captures a wide array of interactions with healthcare facilities, from scheduled appointments and phone consultations to nursing home visits and hospital admissions.

The raw EHR data comprises of a multitude of variables. These include medical diagnostic codes, as categorized by the International Classification of Diseases (ICD-10) and International Classification of Primary Care (ICPC-2), surgical procedure codes \cite{thlproc}, and patient demographic details such as age and gender. Information about each visit, including the specialty of the attending physician, is also incorporated in the data set. To construct a sequential model of patient history, we transformed the tabular data into a sequence of visits. Following this transformation, our dataset comprised 1,050,512 patient sequences spanning 2012 to 2018, with an average of approximately 58 visits per patient.

\subsubsection{MIMIC}

The MIMIC-IV v1.0 dataset \cite{johnson2020mimic,goldberger2000physiobank} comprises EHRs from approximately $250,000$ patients admitted to the Beth Israel Deaconess Medical Center (BIDMC). This extensive dataset includes detailed patient data such as vitals, doctor notes, diagnoses, procedure and medication codes, discharge summaries, and more, collected during both hospital and ICU admissions.

For the purpose of this study, we focused solely on hospital admissions. Relevant patient information was extracted from the {\fontfamily{qcr}\selectfont patients, admissions, diagnoses\_icd, procedures\_icd, drgcodes}, and {\fontfamily{qcr}\selectfont services} tables. Visit information was grouped by the admission identifier {\fontfamily{qcr}\selectfont hadm\_id}, while all visits were grouped by the patient identifier {\fontfamily{qcr}\selectfont subject\_id}. Our modified dataset thus comprised 256,878 patients, averaging 2.04 visits per patient.

\subsection{Prediction tasks}
\label{tasks}

In order to evaluate our model's performance, we established prediction tasks on both the Pummel and MIMIC-IV datasets. For the Pummel dataset, our primary goal was to predict the healthcare utilization of specific patient subgroups, determined by diagnosed diseases. Rather than directly estimating monetary demand, which might vary between countries, we formulated two tasks indicative of healthcare utilization, to serve as a proxy for actual healthcare costs. Using one-year EHR histories, we predicted the following variables for each patient for the subsequent year:

\begin{itemize}
  \item \textbf{Task 1 - Pummel Visits.} The number of physical visits to healthcare centers ($y_{\text{count}}$)
  \item \textbf{Task 2 - Pummel Diagnoses.} The counts of physical visits due to six specific disease categories ($y_{\text{diag}}$)
\end{itemize}

Both of these tasks are intrinsically related to healthcare costs. The Pummel Diagnoses task specifically focuses on six disease categories identified as significant healthcare resource consumers in Finland \cite{thlneedfor}. These include cancer (ICD-10 codes starting with C and some D), endocrine and metabolic diseases (E), diseases related to nervous systems (G), diseases of the circulatory system (I), diseases of the respiratory system (J), and diseases of the digestive system (K).

For both tasks, we modeled the outcome as a Poisson distribution, 
\begin{equation}
    Pr(X=k) = \lambda^k \frac{e^{-k}}{k!}, \quad k = 1, 2, \dots
\end{equation}
and used a neural network to estimate the expected event rate $\lambda$ for each patient. During the training process, the negative log-likelihood of the Poisson distribution is minimized, thereby serving as our loss function.

In the MIMIC dataset:

\begin{itemize}
  \item \textbf{Task 3 - MIMIC Mortality.}. Predicting the probability of inpatient mortality ($y_{\text{death}}$)
\end{itemize}

For MIMIC Mortality task, the {\fontfamily{qcr}\selectfont hospital\_expire\_flag} feature from the MIMIC-IV admissions table is utilized as an indicator of patient mortality during the current hospitalization episode.  To enhance the predictability of this model, the two most recent visits from each patient's history were excluded. As a result, Task 3 effectively estimates the probability of mortality following the subsequent two visits. The loss function employed for this task was binary cross entropy.

\section{Methods}
\label{methods}

Here, we describe our method. Section \ref{pipeline} describes how the raw sequential patient data is converted into an input tensor. Section \ref{sec:components} outlines the components of the SANSformers model: an adaptation of the axial decomposition to decompose the input tensor according to inter-visit and intra-visit dimensions (Section \ref{axial-mixer}), mixers which replace the self-attention to update the token representations based on the decomposed intra-visit and intervisit information (Section \ref{sec:time_mixing}), and details of the positional and temporal encodings (Section \ref{sec:positional_encoding}). Section \ref{sec-pre-train} describes the Generative Summary Pretraining (GSP) approach for self-supervised learning.

\subsection{EHR input transformation pipeline}
\label{pipeline}

Electronic Health Records (EHRs) present a substantial challenge for machine learning applications due to their variable structure, heterogeneous content, and large size. We apply a multi-step transformation to convert the complex raw data into a form suitable for modeling. Let the sequential patient history be represented by $v_{i, t}$, which denote collected visit records for each patient $i$, at time-step $t = 1, 2, \dots, T_i$. Notably, here $t$ is the index of the visit rather than absolute time, and $T_i$ is the total number of visits to a healthcare center for patient $i$. During training, we employ zero-padding to adjust $T_i$ to match the length of the longest sequence within the batch ($T$). consider a batch of patient histories with varying numbers of visits. For example, with a batch size of two patients with 8 and 12 visits, respectively, we pad the shorter sequence with four pad tokens, resulting in sequences of length $T = 12$. 

Our primary task is to predict a patient's healthcare utilization, defined as the number of visits in the year following the first recorded diagnosis of a specific disease. As input, we use one-year patient history leading up to this first diagnosis, what we refer to as the ``fixed time frame''. For example, suppose a patient is diagnosed with Type 2 diabetes on April 5th, 2013. The input for the model in this case would be the sequence of visits from 2012 until the end of 2013. The target that the model is trained to predict would then be the number of hospital visits that the patient made in 2014. Below we elaborate how we form the input tensor from sequential patient history in the fixed time frame.

Each visit record, $v_{i, t}$, in our EHR dataset is characterized by a multivariate sequence of discrete tokens. These tokens signify various types of healthcare interactions such as diagnosis codes, procedure codes, or visit specialties. To accommodate these tokens within our computational framework, we perform a two-step transformation process: numerical encoding followed by projection into a denser vector subspace. First, we map the discrete tokens onto a single shared vocabulary, $\mathcal{W}$, and represent them as one-hot-encoded (OHE) vectors, $\widehat{c} \in \mathbb{R}^{|\mathcal{W}|}$. Subsequently, we project these OHE vectors into dense vectors, $X_{emb} \in \mathbb{R}^{E}$, by multiplying each OHE vector $\widehat{c}$ with an embedding matrix, $W_{emb} \in \mathbb{R}^{|\mathcal{W}| \times E}$. Post-projection, we reshape the patient data to conform to a $T \times V \times E$ structure. Here, $T$ denotes the total number of visits, $V$ represents the number of codes per visit (the intra-visit dimension), and $E$ corresponds to the dimension of the embedding vectors.

Given the variability in patients' visit frequencies, the ``fixed time frame'' approach might yield sparse histories for some patients. To address this, we enrich the temporal representation of visit records by incorporating both the sequence time-step $t$ and the absolute time $\tau$. Additionally, we calculate the time difference $\Delta \tau$ between visits (expressed in days), a strategy that has proven effective in previous studies \cite{choi2016doctor,choi2016retain,li2020behrt,rasmy2021med,ashfaq2019readmission}. The details of the positional encodings are given in Section \ref{sec:positional_encoding}. This approach optimally exploits available temporal data, thereby enhancing the predictive capacity of our model.

\begin{figure}[t]
    \centering
    \includegraphics[width=0.8\linewidth]{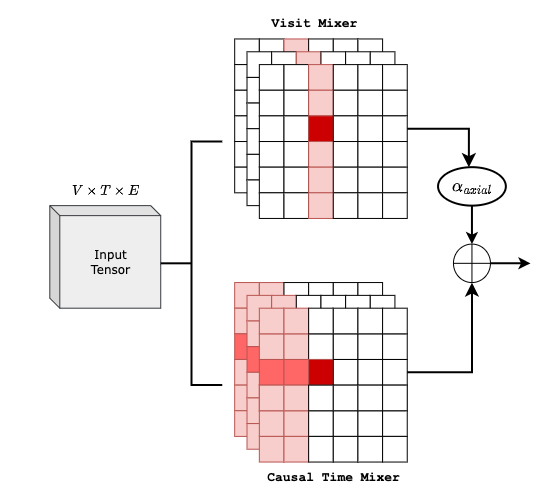}
    \caption{\textbf{Axial Decomposition.} Illustration of the distinct processing of data components in the axial SANSformer model. When updating a specific token (or code) within a visit, the model independently considers all previous visits (through the \textit{Causal Time Mixer}) and all other tokens within the same visit (via the \textit{Visit Mixer}). A prior visit is represented as the summation of its constituent tokens. These separately processed components are combined using a weighted addition mechanism, controlled by the parameter $\alpha_{axial}$, highlighting our approach's adaptability to the unique attributes of each prediction task.}
    \label{fig:axial-mixer}
\end{figure}

\subsection{Components of the SANSformer Model}
\label{sec:components}

\begin{figure*}[t]
  \centering
  \includegraphics[width=\linewidth]{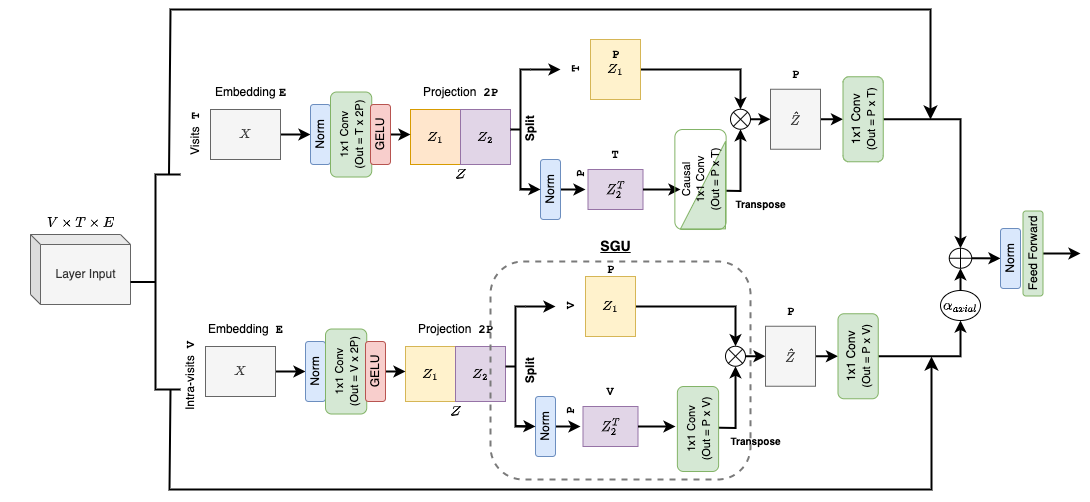}

    \caption[]{\textbf{Detailed schematic of the \textit{axial SANSformer} model.} The original tensor, comprising embedding size (E), intra-visit size (V), and time-steps (T), is axially decomposed to yield two tensors with dimensions \(T \times E\) and \(V \times E\). The model's dataflow bifurcates into two distinct branches: a visit-wise branch (top) and an intra-visit branch (bottom). The visit-wise branch aggregates tokens from preceding visits (enabled by causal masking), while the intra-visit branch focuses on tokens from the current visit. The Spatial Gating Unit (SGU) facilitates the cross-token interactions for each branch. The two branches are integrated using a scalar weight \(\alpha_{axial}\), which is optimized using cross-validation. Consequently, when \(\alpha_{axial}=0\) the model simplifies to the \textit{additive SANSformer} variant.}
  \label{fig:detailed-mixer}
\end{figure*}

The SANSformer maintains several components from the traditional Transformer model, such as positional encodings, skip-connections \cite{he2016deep}, and layer normalization \cite{ba2016layer}, due to their demonstrated efficacy in handling sequential data and aiding model training. While the self-attention mechanism is replaced with the mixing mechanism, preserving other aspects of the Transformer model ensures that SANSformers benefit from the robustness and stability that have been empirically validated in numerous applications. The architecture of one SANSformer layer is shown in Fig. \ref{fig:detailed-mixer}.

\subsubsection{Adapting Axial Decomposition for EHR}
\label{axial-mixer}

Deep learning on Electronic Health Record (EHR) data encounters significant computational challenges when handling multidimensional input sequences. Flattening these sequences along the time-axis and applying traditional attention mechanisms often result in computational complexity of $\mathcal{O}(T^2V^2)$, where $T$ and $V$ represent the time-step and intra-visit size, respectively. We propose an \textit{axial SANSformer} variant that incorporates axial attention, a technique that separately applies mixing along each axis of the input tensor, reducing the complexity to $\mathcal{O}(V^2T + VT^2)$ \cite{ho2019axial}. This mechanism has been optimized for image data, and there the idea is to decompose the attention across the whole image into row-wise and column-wise attention, attending all pixels on the same row/column as the pixel whose representation we are updating.

The axial attention technique for image data requires careful adaptation for the unique structure and semantics of EHR data. In its naive form, axial mixing along the time axis would be equivalent to attending tokens in previous visits at a specific intra-visit index (analogous to attending row-wise pixels). Although meaningful for images, it loses semantic relevance in the EHR context where the order of codes within the previous visits may be arbitrary. We address this issue and modify the axial mixing operation by first representing each visit as the sum of its tokens before applying the mixing, allowing us to capture the comprehensive context of previous visits rather than isolated codes. This modification for Causal Time Mixer is shown in Fig. \ref{fig:axial-mixer}. On the other hand, intra-visit mixer is similar to column-wise mixing for image pixels.

Additionally, we refine the combination of visit-wise and code-wise attended tensors, replacing the simple addition used in the original axial attention mechanism with a weighted average. A parameter, $\alpha_{\text{axial}}$, is introduced and tuned via cross-validation, to balance the contributions of these tensors, catering to the unique requirements of specific prediction tasks. These adaptations enhance the axial SANSformer's effectiveness and interpretability when dealing with complex EHR data. By setting $\alpha_{\text{axial}}$ to zero we get a special case of the model which we call the \textit{additive SANSformer}.

\subsubsection{Introducing Mixers for EHR}
\label{sec:time_mixing}

After the axial decomposition, each patient is represented by two matrices with dimensions $T\times E$ and $V\times E$, corresponding to mixing over time (i.e., across visits) and mixing over tokens (i.e., intra-visit), respectively. The SANSformer then performs a non-linear operation that transforms an input tensor $x_{in} \in \mathbb{R}^B \times \mathbb{R}^T \times \mathbb{R}^E$, where $B$ is the batch size, $T$ the sequence length and $E$ the embedding dimension, onto an output tensor of the same dimensions. Here, we describe only the mixing over time, corresponding to the upper branch in Fig. \ref{fig:detailed-mixer}; mixing over tokens is similar except for the causal masking (lower branch in Fig. \ref{fig:detailed-mixer}).

In Transformer models, cross-time interaction crucial, and conventionally achieved by self-attention mechanisms. However, recent research \cite{tolstikhin2021mlp,liu2021pay,touvron2021resmlp} has shown that similar results can be obtained by implementing feedforward layers across the time axis, a process we refer to as ``mixers''. These facilitate cross-token interaction along a specific axis through matrix multiplication, functioning similarly to $1 \times 1$ convolutions where the number of channels is equal to the size of the hidden dimension. Despite capturing only second-order interactions, compared to the third-order interactions encompassed by self-attention, see \cite{liu2021pay}, we hypothesize that ``mixers'' provide a satisfactory level of interaction for most applications in the realm of EHRs.

Our method of achieving cross-token interaction through mixing draws inspiration from the Spatial Gating Unit (SGU) introduced in \cite{liu2021pay}. Given an input $X \in \mathbb{R}^{T \times E}$ with sequence length $T$ and embedding dimension $E$, we transform it into an output $Y$ of identical dimensions. The transformation process is detailed as follows:
\begin{align}
    Z           &= \text{GELU}(X U) \\
    \label{eq:sgu} \widehat{Z} &= \text{SGU}(Z) \\
    Y           &= \widehat{Z}V
\end{align}
Here, $U \in \mathbb{R}^{E \times 2P}$ and $V \in \mathbb{R}^{P \times E}$ are trainable weight matrices, while $P$ denotes the projection dimension, usually larger than $E$. We use Gaussian Error Linear Units (GELU) \cite{hendrycks2016gaussian}, a preferred activation function in modern Transformer models, including BERT.

In Equation (\ref{eq:sgu}), the SGU function carries out cross-time mixing by dividing $Z \in \mathbb{R}^{T \times 2P}$ along the projected dimension into two portions, $Z_1, Z_2, \in \mathbb{R}^{T \times P}$. An affine transformation is then applied to one of these segments:
\begin{align}
    Z_1, Z_2            &= \text{split}(Z) \\
    \label{eq:sguW} \widehat{Z_{2}}^T   &= \text{GELU}(Z_2^T W + b) \\
    \widehat{Z}         &= Z_1 \odot \widehat{Z_{2}}
\end{align}
Here, $W \in \mathbb{R}^{T \times T}$ is another trainable weight matrix, and $\odot$ denotes element-wise multiplication. For simplicity, normalization operations and skip-connections have been left out from the equations above. The GELU activation is used in all mixer components.
To facilitate autoregressive model training, we introduce causal masking on the SGU weight matrix, $W$, to prevent future time-steps from leaking information. This is efficiently done by zeroing all upper-triangular elements of the matrix before the multiplication in (\ref{eq:sguW}).

\subsubsection{Incorporating Positional and Temporal Encodings}
\label{sec:positional_encoding}

While SANSformers are effective at learning complex patterns, they are intrinsically order-invariant, which means they do not inherently consider the temporal sequence of tokens. To integrate this sequence information, we use positional encodings, which are implemented using sinusoidal functions with varying wavelengths. These encode different positions along the embedding axis, providing unique representations for each token's position. $\text{PE}(t, 2i) = \sin(t/10000^{2i/d_{model}})$ and $\text{PE}(t, 2i+1) = \cos(t/10000^{2i/d_{model}})$. Here, $\text{PE}$ represents the positional encoding for position $t$ and dimension $2i$ or $2i+1$. The variable $d_{model}$ denotes the embedding dimensionality, while $t$ and $i$ denote the position and dimension respectively.

While these encodings uniquely identify positions, i.e., visit indices, they are insufficient for EHR data due to temporal aspects. Specifically, the relevance of two visits typically diminishes as the time gap between them increases. To address this, we introduce $\Delta \tau$ embeddings, which represent the elapsed days between consecutive visits. Consequently, the enhanced input to the model is the element-wise addition of the original input, positional encodings, and $\Delta \tau$ encodings. This approach incorporates both static and dynamic elements from a patient's medical history, providing a more comprehensive representation for the model.

\subsection{Generative Summary Pre-training (GSP)}
\label{sec-pre-train}

Our focus on divergent patient subgroups, defined as patients with a certain rare disease diagnosis, necessitates a consideration of patient history only up until the first incidence of the given diagnosis. This context imposes constraints on the data accessible for predictive tasks, in terms of both the size of disease-specific subgroups and the temporal data available. Furthermore, our task confronts an additional challenge often faced in healthcare resource allocation: predictions for a specific year typically rely on just the preceding year's history. Factors contributing to this practice include limited patient histories, patients moving and changing healthcare providers, and the annual cycle of resource allocation \cite{ellis2018risk}. However, we note that the Pummel dataset, with its seven-year history per patient, has previously demonstrated potential to enhance predictive accuracy further by using longer histories when available \cite{kumar2020predicting}.

To optimize the use of available data and address these challenges, we propose the Generative Summary Pre-training (GSP) method. GSP is a self-supervised pre-training strategy designed to utilize the abundant patient data in the general population outside the target patient subgroups. Furthermore, it would be possible to capitalize on the temporal data in the years before the first diagnosis for the target patients that is otherwise discarded. However, to maintain the integrity of our model's predictions and prevent information leakage between the pre-training and fine-tuning phases, we restrict in this work the pre-training phase to patients not part of any specific subgroup considered in our fine-tuning tasks. This strategy ensures a clear separation of data between the two phases, fortifying the reliability of our approach. The training objective for GSP is to generate a summary prediction of healthcare utilization for the upcoming years in a sequential fashion (in the general population, i.e., outside the target subgroups), based on the health records of the preceding year - a target that aligns closely with our primary task.

\section{Results}

\begin{table*}[ht]

    \caption{\textbf{Performance on Pummel Visits task.} Performance comparison of models on the Pummel test set for Type 2 diabetes subgroup ($N$=41,761), without pretraining. Mean performance $\pm$ standard deviation from five random restarts is reported.}

    \label{ri-results-task1}
    \centering
    \small
        \centering
        \begin{tabular}{@{}lccc@{}}
            \toprule
            \multicolumn{1}{c}{\textbf{Model}}  & \textbf{\makecell{$R^2$ \textuparrow \\ $y_{\text{count}}$}} & \textbf{\makecell{Spearman \\ Corr. \textuparrow \\ $y_{\text{count}}$}} & \textbf{\makecell{MAE \textdownarrow\\ $y_{\text{count}}$}} \\ \midrule
            Lasso                           & 0.121   & 0.432   & 6.394   \\
            RETAIN \cite{choi2016retain}    & 0.298 $\pm$ 0.003     & 0.509 $\pm$ 0.003   & 5.474 $\pm$ 0.081   \\
            BEHRT \cite{li2020behrt}        & 0.272 $\pm$ 0.003     & 0.497 $\pm$ 0.001 & 5.654 $\pm$ 0.069 \\
            BRLTM \cite{meng2021bidirectional} & 0.276 $\pm$ 0.003 & 0.503 $\pm$ 0.001 & 5.600 $\pm$ 0.027 \\
            SARD \cite{kodialam2021deep}    & 0.295 $\pm$ 0.007   & 0.516 $\pm$ 0.000   & 5.547 $\pm$ 0.110 \\
            \midrule
            Additive SANSformer (ours)      & \textbf{0.311} $\pm$ \textbf{0.002}   & \textbf{0.521} $\pm$ \textbf{0.003}   & \textbf{5.413} $\pm$ \textbf{0.024}   \\
            Axial SANSformer (ours)         & 0.306 $\pm$ 0.003   & 0.518 $\pm$ 0.003   & 5.453 $\pm$ 0.010 \\ \bottomrule        
    \end{tabular}
\end{table*}

\begin{table*}[ht]
\begin{minipage}{.48\textwidth}
    \caption{\textbf{Performance on Pummel Diagnoses task.} Performance comparison on the Pummel test set for counts of visits due to six disease categories, without pretraining. Mean metrics from six categories are reported. For individual scores, see appendix.}
\label{ri-results-task2}
\centering
\small
        \centering
        \begin{tabular}{@{}lcc@{}}
        \toprule
        \multicolumn{1}{c}{\textbf{Model}}  & \textbf{\makecell{Spearman \textuparrow \\ Corr. $y_{\text{diag}}$}} & \textbf{\makecell{MAE \textdownarrow\\ $y_{\text{diag}}$}} \\ \midrule
        Lasso                           & 0.200    & 1.212  \\
        RETAIN \cite{choi2016retain}    & 0.249 $\pm$ 0.008 & 1.093 $\pm$ 0.045 \\
        BEHRT \cite{li2020behrt}        & 0.225 $\pm$ 0.005 & 1.122 $\pm$ 0.009 \\
        BRLTM \cite{meng2021bidirectional}    & 0.259 $\pm$ 0.004 & \textbf{1.083 $\pm$ 0.016} \\
        SARD \cite{kodialam2021deep}    & 0.233 $\pm$ 0.024 & 1.111 $\pm$ 0.095 \\
        \midrule
        Additive SANSformer (ours)      & 0.260 $\pm$ 0.004 & 1.123 $\pm$ 0.016 \\
        Axial SANSformer (ours)         & \textbf{0.262 $\pm$ 0.006} & 1.117 $\pm$ 0.014 \\ \bottomrule
        \end{tabular}
        \end{minipage}
\hspace{.04\textwidth}%
\begin{minipage}{.48\textwidth}
    \caption{\textbf{Performance on MIMIC Mortality task.} Model comparison on the MIMIC-IV dataset for the in-hospital mortality task, with mean AUC score $\pm$ standard deviation from five random restarts.}
    \label{ri-results-task3}
    \small
    \centering
    \begin{tabular}{l c}
    \toprule
    \multicolumn{1}{c}{\textbf{Model}} & \textbf{AUC \textuparrow} \\
    \midrule
    $L_1$-reg. Logistic Regression \cite{razavian2015population}      & 0.728 \\
    RETAIN \cite{choi2016retain}         & 0.707 $\pm$ 0.007 \\
    BEHRT \cite{li2020behrt}            & 0.693 $\pm$ 0.006 \\
    BRLTM \cite{meng2021bidirectional}            & 0.695 $\pm$ 0.004 \\
    SARD \cite{kodialam2021deep}         & 0.742 $\pm$ 0.002\\
    \midrule
    Additive SANSformer (ours)         & 0.759 $\pm$ 0.002\\
    Axial SANSformer (ours)            & \textbf{0.761} $\pm$ \textbf{0.004}\\
    \bottomrule
    \end{tabular}
    \end{minipage}
\end{table*}

\begin{table*}[bt]
    \caption{\textbf{Performance on Divergent Subgroups in Pummel Dataset.} The table compares performance in Bipolar disorder (N=827) and Multiple Sclerosis (N=123) subgroups. In each table, the results from the randomly initialized models are shown on top and the corresponding results from models pretrained using Generative Summary Pre-training (GSP), Masked Language Model (MLM) and Reverse Distillation (RD) \cite{kodialam2021deep} are shown below. Mean performance $\pm$ standard deviation from five random restarts is reported. }
    \label{divergent-results}
    \centering
    \small

    \begin{tabular}{lccc|c}
    \toprule
        \textbf{Bipolar disorder (N=827)} & \multicolumn{2}{c|}{\textbf{Pummel Visits}} & \multicolumn{2}{c}{\textbf{Pummel Diagnoses}} \\ \midrule
        
        \multicolumn{1}{c}{\textbf{Model}} & \textbf{\makecell{Spearman \textuparrow\\ Corr. $y_{\text{count}}$}} & \textbf{\makecell{MAE \textdownarrow\\ $y_{\text{count}}$}} & \textbf{\makecell{Spearman \textuparrow \\ Corr. $y_{\text{diag}}$}} & \textbf{\makecell{MAE \textdownarrow \\ $y_{\text{diag}}$}} \\ \midrule
        Lasso                               & 0.446 & \multicolumn{1}{c|}{8.575} & 0.277 & 1.311 \\ 
        RETAIN \cite{choi2016retain}         & 0.112 $\pm$ 0.207 & \multicolumn{1}{c|}{9.854 $\pm$ 0.110}  & 0.029 $\pm$ 0.033 & 1.535 $\pm$ 0.035  \\
        BEHRT \cite{li2020behrt}   & 0.117 $\pm$ 0.020 & \multicolumn{1}{c|}{9.815 $\pm$ 0.143} & 0.101 $\pm$ 0.014 & 1.446 $\pm$ 0.019 \\
        BRLTM \cite{meng2021bidirectional}   & 0.124 $\pm$ 0.014 & \multicolumn{1}{c|}{9.840 $\pm$ 0.143} & 0.115 $\pm$ 0.005 & 1.441 $\pm$ 0.014 \\
        SARD \cite{kodialam2021deep}         & 0.136 $\pm$ 0.006 & \multicolumn{1}{c|}{9.294 $\pm$ 0.050} & 0.012 $\pm$ 0.021 & 1.442 $\pm$ 0.028 \\
        Additive SANSformer                 & 0.092 $\pm$ 0.014 & \multicolumn{1}{c|}{9.787 $\pm$ 0.086} & 0.077 $\pm$ 0.028 & 1.493 $\pm$ 0.013\\
        Axial SANSformer                    & 0.091 $\pm$ 0.074 & \multicolumn{1}{c|}{10.062 $\pm$ 0.122} & 0.077 $\pm$ 0.043 & 1.671 $\pm$ 0.127   \\
        \midrule
        RETAIN + GSP                        & 0.398 $\pm$ 0.010 & \multicolumn{1}{c|}{8.466 $\pm$ 0.025} & 0.322 $\pm$ 0.014 & 1.249 $\pm$ 0.014  \\
        BEHRT + MLM                          & 0.117 $\pm$ 0.020 & \multicolumn{1}{c|}{9.815 $\pm$ 0.143} & 0.279 $\pm$ 0.011 & 1.382 $\pm$ 0.068  \\
        BRLTM + MLM                           & 0.401 $\pm$ 0.020 & \multicolumn{1}{c|}{8.685 $\pm$ 0.307} & 0.298 $\pm$ 0.017 & 1.340 $\pm$ 0.056  \\
        SARD + RD                           & 0.439 $\pm$ 0.006 & \multicolumn{1}{c|}{8.372 $\pm$ 0.085} & 0.047 $\pm$ 0.008 & 1.463 $\pm$ 0.145  \\
        SARD + GSP                          & 0.442 $\pm$ 0.016 & \multicolumn{1}{c|}{8.375 $\pm$ 0.125} & 0.305 $\pm$ 0.008 & 1.453 $\pm$ 0.270  \\
        Additive SANSformer + GSP           & \textbf{0.548 $\pm$ 0.006} & \multicolumn{1}{c|}{8.085 $\pm$ 0.205} & 0.181 $\pm$ 0.007 & 1.463 $\pm$ 0.009  \\
        Axial SANSformer + GSP              & 0.522 $\pm$ 0.018 & \multicolumn{1}{c|}{\textbf{7.709 $\pm$ 0.138}} & \textbf{0.372 $\pm$ 0.009} & \textbf{1.125 $\pm$ 0.034} \\    
        \bottomrule
    \end{tabular}

\bigbreak

    \begin{tabular}{lccc|c}
    \toprule
        \textbf{Multiple Sclerosis (N=123)} & \multicolumn{2}{c|}{\textbf{Pummel Visits}} & \multicolumn{2}{c}{\textbf{Pummel Diagnoses}} \\ \midrule
        
        \multicolumn{1}{c}{\textbf{Model}}  & \textbf{\makecell{Spearman \textuparrow \\ Corr. $y_{\text{count}}$}} & \textbf{\makecell{MAE \textdownarrow \\ $y_{\text{count}}$}} & \textbf{\makecell{Spearman \textuparrow \\ Corr. $y_{\text{diag}}$}} & \textbf{\makecell{MAE \textdownarrow \\ $y_{\text{diag}}$}} \\ \midrule
        Lasso                               & \textbf{0.533} & \multicolumn{1}{c|}{8.413} & 0.160 & 1.119 \\
        RETAIN \cite{choi2016retain}         & 0.088 $\pm$ 0.133 & \multicolumn{1}{c|}{8.800 $\pm$ 0.316}  & 0.061 $\pm$ 0.045 & 1.288 $\pm$ 0.029  \\
        BEHRT \cite{li2020behrt}   & 0.010 $\pm$ 0.075 & \multicolumn{1}{c|}{9.781 $\pm$ 0.534} & 0.026 $\pm$ 0.080 & 1.250 $\pm$ 0.028 \\
        BRLTM \cite{meng2021bidirectional}   & 0.013 $\pm$ 0.074 & \multicolumn{1}{c|}{9.776 $\pm$ 0.541} & 0.036 $\pm$ 0.073 & 1.253 $\pm$ 0.028 \\
        SARD \cite{kodialam2021deep}         & 0.017 $\pm$ 0.046 & \multicolumn{1}{c|}{8.670 $\pm$ 0.063} & 0.010 $\pm$ 0.089 & 1.522 $\pm$ 0.043 \\
        Additive SANSformer                 & 0.019 $\pm$ 0.107 & \multicolumn{1}{c|}{8.983 $\pm$ 0.323} & -0.073 $\pm$ 0.041 & 1.328 $\pm$ 0.063\\
        Axial SANSformer                    & 0.174 $\pm$ 0.196 & \multicolumn{1}{c|}{8.879 $\pm$ 0.146} & 0.000 $\pm$ 0.059 & 1.300 $\pm$ 0.059   \\
        \midrule
        
        RETAIN + GSP                        & 0.358 $\pm$ 0.031 & \multicolumn{1}{c|}{9.056 $\pm$ 0.130} & 0.230 $\pm$ 0.011 & 1.069 $\pm$ 0.026  \\
        BEHRT + MLM                           & 0.263 $\pm$ 0.094 & \multicolumn{1}{c|}{8.031 $\pm$ 0.727} & 0.136 $\pm$ 0.036 & 1.112 $\pm$ 0.023  \\
        BRLTM + MLM                           & 0.316 $\pm$ 0.126 & \multicolumn{1}{c|}{7.760 $\pm$ 0.580} & 0.159 $\pm$ 0.033 & 1.110 $\pm$ 0.015  \\
        SARD + RD                           & 0.082 $\pm$ 0.049 & \multicolumn{1}{c|}{8.610 $\pm$ 0.151} & 0.013 $\pm$ 0.016 & 1.518 $\pm$ 0.052  \\
        SARD + GSP                          & 0.343 $\pm$ 0.036 & \multicolumn{1}{c|}{\textbf{7.711 $\pm$ 0.064}} & 0.239 $\pm$ 0.015 & \textbf{1.034 $\pm$ 0.017}  \\
        Additive SANSformer + GSP           & 0.328 $\pm$ 0.018 & \multicolumn{1}{c|}{8.104 $\pm$ 0.186} & 0.275 $\pm$ 0.042  & 1.046 $\pm$ 0.028    \\
        Axial SANSformer + GSP              & 0.510 $\pm$ 0.017 & \multicolumn{1}{c|}{7.783 $\pm$ 0.227} & \textbf{0.288 $\pm$ 0.012} & 1.063 $\pm$ 0.028 \\    
        \bottomrule
    \end{tabular}
\end{table*}

In this section, we evaluate the performance of the SANSformer model across the tasks outlined in Section \ref{tasks}, demonstrating its ability to handle the complexities of EHR data effectively. Our primary objective is to examine its predictive capabilities across various patient subgroups. Through this comprehensive performance analysis, we aim to showcase the model's predictive strength, scalability, and applicability in real-world healthcare contexts.

First, we compare our proposed models against several established baselines without any pre-training. The baselines include the RETAIN model \cite{choi2016retain}, which leverages Recurrent Neural Networks (RNNs) for EHR, the SARD model \cite{kodialam2021deep}, a state-of-the-art Transformer-based model crafted for clinical code sequences, and the traditional Lasso/Logistic regression models, serving as linear benchmarks. These comparisons are conducted on both Pummel and MIMIC datasets.

\begin{figure*}[t]
  \centering
  \includegraphics[width=0.85\linewidth]{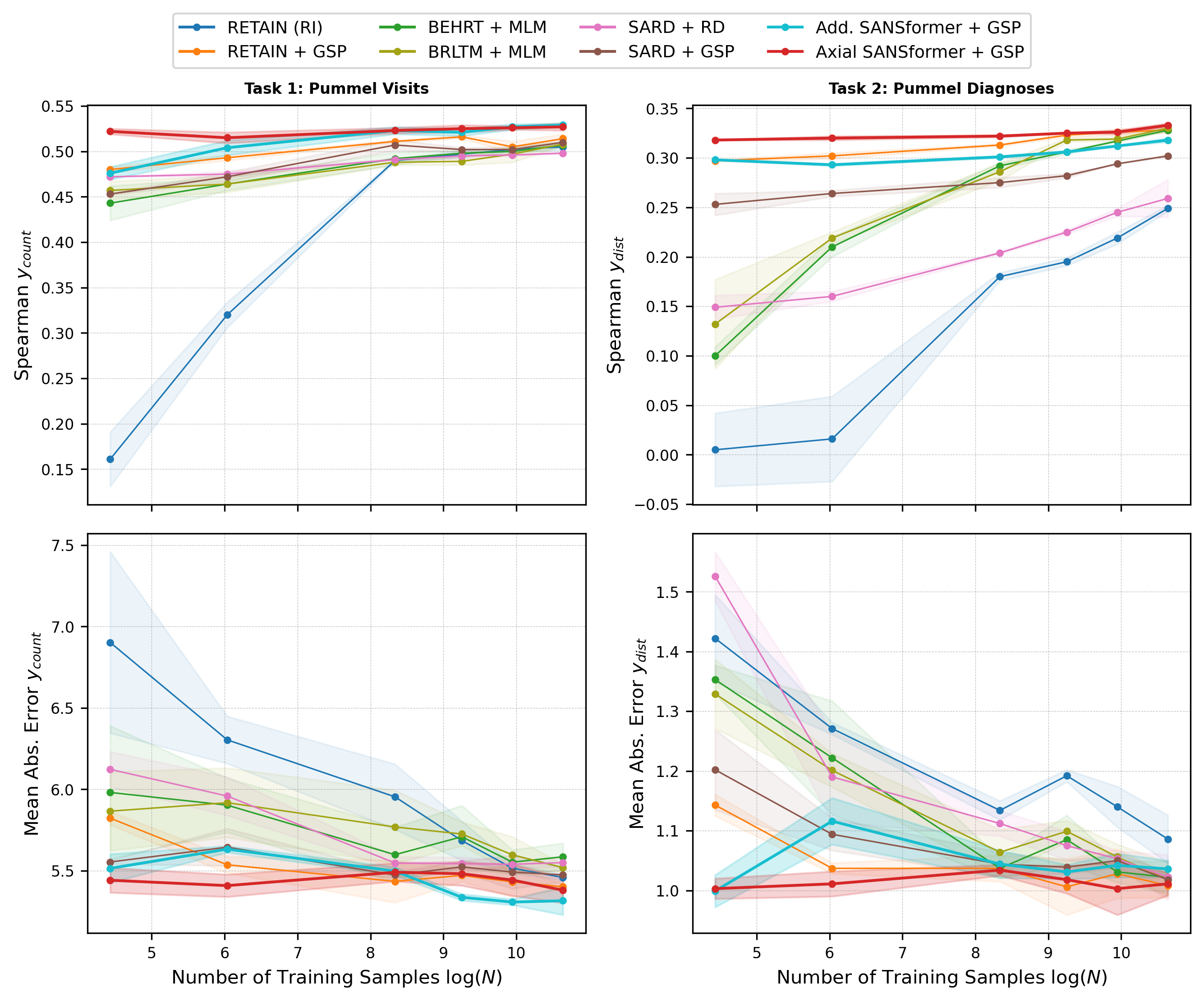}
    \caption[]{\textbf{Data Efficiency through Pretraining on the Pummel Dataset (Type 2 Diabetes Subgroup).} The figure evaluates the performance of pretrained models across varying training data sizes. \emph{Row 1} depicts the Spearman's rank correlation for Pummel Visits and Pummel Diagnoses tasks, while \emph{Row 2} reports the Mean Absolute Error (MAE) for the same tasks. For each data point, the mean and standard deviation computed from three random restarts are shown. The performance of the randomly initialized RETAIN (RI) model has been included for reference.}
    \label{fig:data-ablation-pt}
\end{figure*}

Following this, we demonstrate the advantages offered by our Generative Summary Pre-training (GSP) strategy, studying its potential to improve prediction accuracy in three divergent subgroups within the Pummel dataset: patients diagnosed with Type 2 Diabetes, Bipolar Disorder, or Multiple Sclerosis. Due to the larger size of the diabetes subgroup in Pummel, we repeat the analysis with varying training data sizes, allowing us to evaluate data efficiency, as illustrated in Fig. \ref{fig:data-ablation-pt}. Finally, we assess the model's performance on the two smaller subgroups, which provides insight into how effectively our model scales and adjusts to the challenging scenario of very small cohort sizes.

\subsection{Details of the Experiments}

For our study, we used patients diagnosed with the specific disease between the years 2012 - 2015 as the training set, while those diagnosed in 2016 formed the test set. The training data is further divided, assigning 80\% for model training and the remaining 20\% for validation. Our baseline models, which include L1 Regularized Logistic Regression, RETAIN \cite{choi2016retain}, and SARD \cite{kodialam2021deep}, are compared with our proposed axial and additive SANSformer variants. We average the hidden tokens from each timestep to obtain the logits for SARD on regression tasks.

All models are implemented using PyTorch \cite{paszke2017automatic}, and trained with the Rectified Adam optimizer \cite{liu2019variance} with a cyclical learning rate schedule \cite{smith2017cyclical} and linear decay. The computation was performed using a single NVIDIA Tesla V100 GPU. The models were trained for 20 epochs with a batch size of 32. For the experiments with pretrained model on the smaller subgroups, we reduced the batch size to 8. Both axial and additive SANSformer variants are included in the comparisons, each composed of four layers and utilizing 16 attention heads. All SANSformer models have a fixed embedding dimension of 256, and are subject to gradient clipping at $10$ to prevent gradient explosions. The model hyperparameters were tuned using Optuna \cite{optuna_2019} on the validation set and the search ranges are provided in the Supplement section.

The performance of our models on the PUMMEL dataset is evaluated using Spearman's rank correlation and Mean Absolute Error (MAE) for visit count predictions (Task 1), metrics widely used for example in the risk adjustment literature \cite{Layton2018EvaluatingTP, sheen2019metastasis, cueto2019comparative}. For Pummel Diagnoses (Task 2), we present metrics averaged across six disease category counts. For the MIMIC dataset, the performance of mortality prediction (Task 3) is reported using the Area Under the Receiver Operating Characteristic Curve (AUC).

\subsection{Baseline Comparisons without Pretraining}
\label{sec:results-ri}

In this section, we examine the effectiveness of the proposed SANSformer model against the selected baseline models without pretraining. This provides an assessment of our models in a controlled setup, where each model is trained from scratch for the task at hand, minimizing external influences.

The results from the Pummel dataset (from the diabetes subgroup consisting of approximately 41,000 training samples) and the MIMIC dataset, corresponding to Pummel Visits, Pummel Diagnoses and MIMIC Mortality tasks, are presented in Tables \ref{ri-results-task1}, \ref{ri-results-task2} and \ref{ri-results-task3}. By isolating the effects of pretraining, this comparison reveals the inherent capabilities of the SANSformer model in processing EHR data and provides a preliminary insight into its predictive performance.

A close examination of the results shows that our SANSformer model consistently outperforms the baseline models across tasks and metrics, except for the MAE metric in Task 2, where RETAIN is slightly better. This is a strong indication of the model's potential, confirming that the inductive biases we have integrated into it contribute significantly to its performance in handling EHR data. These foundational results establish the context for our main experiment on the larger PUMMEL dataset, leveraging the generative summary pretraining strategy.

\subsection{Transfer Learning for Divergent Subgroup Prediction}
\label{sec:resutls-transfer-learn}

We next investigate the transfer learning capabilities of the SANSformer model when applied to three distinct subgroups in the PUMMEL dataset: Type 2 Diabetes (T2D, N=41,000), Bipolar Disorder (BP, N=800), and Multiple Sclerosis (MS, N=120). The model is first pretrained, facilitated by Generative Summary Pretraining (GSP), on the broader population data (N$\approx$1 million), following which it is fine-tuned on these individual subgroups. This methodology permits us to evaluate the model's adaptability and effectiveness in transfer learning and its performance on Pummel Visits and Pummel Diagnoses tasks, as detailed in Section \ref{tasks}. 

Data efficiency is evaluated by manipulating the size of the training dataset from the T2D subgroup, which provides a sufficiently large sample. This approach addresses the challenge that reduced neural network performance is often associated with smaller training data sizes \cite{Goodfellow-et-al-2016}. Fig. \ref{fig:data-ablation-pt} presents the results from pre-trained models utilizing various data sizes from the T2D subgroup. The results demonstrate that our axial SANSformer model is the most data efficient, providing superior performance in both Pummel Visits and Pummel Diagnoses tasks.

We further contrast the performance of GSP-pretrained models with models that have been randomly initialized. This comparison is conducted on the MS and BP subgroups, which are smaller and present a dual challenge due to their size and highly skewed target histograms (as illustrated in Fig. \ref{fig:div-subgroups}). As shown in Table \ref{divergent-results}, GSP provides a significant boost in results across all model architectures when compared to random initialization. In these smaller, divergent subgroups, the axial SANSformer model consistently either outperforms or matches the best performing models, highlighting the robustness and versatility of GSP in handling diverse and complex tasks.

Examining challenging scenarios, such as the extremely small and skewed Multiple Sclerosis (MS) subgroup (N=123), brings out the robustness of the Generative Summary Pretraining (GSP) strategy. Even under such conditions, models pretrained with GSP, including our axial SANSformer, retain competitive performance. This emphasizes the value of GSP in maintaining model effectiveness across diverse situations, as opposed to relying on random initialization or other strategies like reverse distillation.

\begin{figure}[]
  \centering
  \includegraphics[scale=0.65]{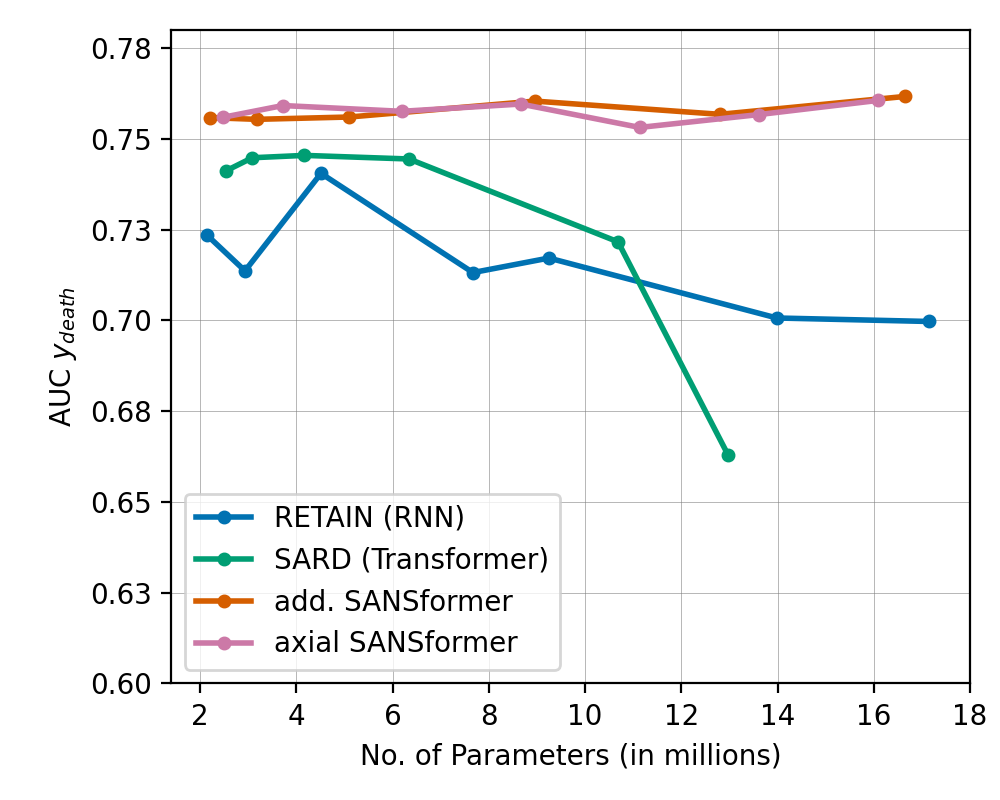}
  \caption[]{\textbf{Parameter Scalability on MIMIC.} This figure plots model performance on Task 3 against the number of trainable parameters, illustrating the scalability of each model. Impressively, both the additive and axial SANSformer models maintain strong performance even as the number of parameters is increased. This is particularly noteworthy given the constant training data size, as the SANSformer models resist overfitting and showcase their superior generalization capacity.}
  \label{fig:mimic-param}
\end{figure}

\subsection{Scalability in SANSformers}
\label{sec:results-scalability}

A significant advantage of Transformer models is their scalability, the property that allows enhanced performance as model size increases. This trait has led to the development of larger Transformer models that redefine the state-of-the-art in many fields\cite{shoeybi2019megatron,fedus2021switch,openai2023gpt4}. To evaluate the scalability of our SANSformers, we conduct experiments on the MIMIC dataset with Task 3, systematically increasing the number of parameters in the models.

As depicted in Fig. \ref{fig:mimic-param}, both additive and axial versions of SANSformer exhibit scalability on par with traditional Transformers when the number of parameters is increased. Throughout these experiments, the size of our training data remains unchanged. As a result, larger models could tend towards overfitting, as was noted in the RETAIN and SARD baselines. However, even with their increased parameter count, SANSformers resist this overfitting trend, signifying their robustness and suggesting enhanced generalization capabilities. These findings underscore the potential of SANSformers as a potent tool in processing and interpreting EHR data.

\subsubsection*{Summary of Results} Throughout our comprehensive evaluations, the SANSformer models consistently exhibit superior performance. They either surpass or match closely with the baselines in the non-pretraining experiments, emphasizing the value of the integrated inductive biases, such as axial decomposition and $\Delta \tau$ embeddings. A detailed ablation study on their effectiveness across both Pummel and MIMIC datasets is presented in Section \ref{sec: ablation}. Furthermore, in the divergent subgroup analysis, neural network models derive substantial benefits from an initial pretraining step. Notably, the introduction of the GSP pretraining strategy enhances model performance, achieving gains beyond other strategies like MLM and RD.

\section{Conclusion}

Our study presents attention-free MLP models \cite{tolstikhin2021mlp,liu2021pay} in the EHR domain, which have previously demonstrated competitive performance in computer vision and NLP tasks. The proposed SANSformer model, in particular, exhibited robust performance across various datasets and tasks. Regardless of whether initialized randomly or pre-trained, it consistently outperformed other strong baselines, thereby consolidating its advantages as discussed in Sections \ref{sec:results-ri}, \ref{sec:resutls-transfer-learn} \& \ref{sec:results-scalability}.

In predicting for divergent subgroups with pre-training, SANSformers consistently match or outperform the other competitive baselines, as depicted in Fig. \ref{fig:data-ablation-pt} and Table \ref{divergent-results}. The advantages of axial mixing over additive visit summarization are most evident in tasks that require the capture of complex interactions between tokens, such as Task 2 in Table \ref{divergent-results} and Fig. \ref{fig:data-ablation-pt}. Without pre-training, SANSformer models consistently outperform other baselines even when randomly initialized, as seen in Tables \ref{ri-results-task1}, \ref{ri-results-task2} and \ref{ri-results-task3}. This highlights that SANSformer models are more parameter-efficient than Transformers while maintaining their scalability, as illustrated in Fig. \ref{fig:mimic-param}.

These results underline the substantial potential of self-supervised pre-training on the general population to amplify prediction accuracy across model architectures in divergent subgroups, providing a boost to subgroup-specific prediction models that are essential in allocating valuable healthcare resources. Interpreting the features learned by SANSformer models and comparing those with features engineered by domain experts could be an interesting extension of this work.

\textbf{Limitations. }Our proposed SANSformer models introduce a notable advancement in utilizing attention-free MLP models within the EHR domain. However, there are inherent limitations to be considered. Firstly, SANSformers encounted a restriction related to sequence length, defined by the dimensions of the weight matrix $W$ (refer to (\ref{eq:sguW}). However, such a limitation is also observed in RNNs and Transformers, where, despite their theoretical capability to process indefinitely long sequences, practical challenges such as the `vanishing gradient' problem and memory limitations, respectively, inhibit their effectiveness for processing of lengthy inputs. Secondly, while our model effectively utilizes inpatient encounters as an indicator to anticipate healthcare utilization, it does not directly compute costs required by some applications \cite{rose2016machine}, relying instead on a proxy. Consequently, further adaptations and enhancements would be required to derive the actual expected healthcare cost. Lastly, the exclusion of the self-attention mechanism from SANSformers ostensibly reduces the model's interpretability. While attention maps, from the self-attention mechanism, provide a degree of insight into model explainability, the clarity and reliability of such interpretability are subject of ongoing debate within the research community \cite{jain2019attention, wiegreffe2019attention}. Applying alternative model-agnostic interpretability methods, such as SHAP \cite{lundberg2017unified}, to furnish insightful and trustworthy explanations of SANSformer's decision making process in healthcare contexts is an exciting direction for future work.

\newpage
\section*{Acknowledgment}

The authors wish to acknowledge CSC – IT Center for Science, Finland, for computational resources, and the computational resources provided by the Aalto Science-IT project. For grammatical coherence and LaTeX formatting, OpenAI's ChatGPT was used to assist in polishing the language and structure in the manuscript.

\bibliographystyle{plain}
\bibliography{pummel-transfer}

\begin{thebibliography}{10}

\bibitem{ahmad2018death}
Muhammad Ahmad, Carly Eckert, Greg McKelvey, Kiyana Zolfagar, Anam Zahid, and
  Ankur Teredesai.
\newblock Death vs. {D}ata {S}cience: {P}redicting end of life.
\newblock In {\em Proceedings of the AAAI Conference on Artificial
  Intelligence}, volume~32, 2018.

\bibitem{optuna_2019}
Takuya Akiba, Shotaro Sano, Toshihiko Yanase, Takeru Ohta, and Masanori Koyama.
\newblock Optuna: A next-generation hyperparameter optimization framework.
\newblock In {\em Proceedings of the 25rd {ACM} {SIGKDD} International
  Conference on Knowledge Discovery and Data Mining}, 2019.

\bibitem{ashfaq2019readmission}
Awais Ashfaq, Anita Sant’Anna, Markus Lingman, and S{\l}awomir Nowaczyk.
\newblock Readmission prediction using deep learning on electronic health
  records.
\newblock {\em Journal of Biomedical Informatics}, 97:103256, 2019.

\bibitem{avati2018improving}
Anand Avati, Kenneth Jung, Stephanie Harman, Lance Downing, Andrew Ng, and
  Nigam~H Shah.
\newblock Improving palliative care with deep learning.
\newblock {\em {BMC} medical informatics and decision making}, 18(4):55--64,
  2018.

\bibitem{ba2016layer}
Jimmy~Lei Ba, Jamie~Ryan Kiros, and Geoffrey~E Hinton.
\newblock Layer normalization.
\newblock {\em arXiv preprint arXiv:1607.06450}, 2016.

\bibitem{bellamy2020evaluating}
David Bellamy, Leo Celi, and Andrew~L Beam.
\newblock Evaluating progress on machine learning for longitudinal electronic
  healthcare data.
\newblock {\em arXiv preprint arXiv:2010.01149}, 2020.

\bibitem{pmlr-v27-bengio12a}
Yoshua Bengio.
\newblock Deep learning of representations for unsupervised and transfer
  learning.
\newblock In Isabelle Guyon, Gideon Dror, Vincent Lemaire, Graham Taylor, and
  Daniel Silver, editors, {\em Proceedings of {ICML} {W}orkshop on
  {U}nsupervised and {T}ransfer {L}earning}, volume~27 of {\em Proceedings of
  Machine Learning Research}, pages 17--36, Bellevue, Washington, USA, 02 Jul
  2012. PMLR.

\bibitem{brown2020language}
Tom~B Brown, Benjamin Mann, Nick Ryder, Melanie Subbiah, Jared Kaplan, Prafulla
  Dhariwal, Arvind Neelakantan, Pranav Shyam, Girish Sastry, Amanda Askell,
  et~al.
\newblock Language models are few-shot learners.
\newblock {\em arXiv preprint arXiv:2005.14165}, 2020.

\bibitem{chen2020generative}
Mark Chen, Alec Radford, Rewon Child, Jeffrey Wu, Heewoo Jun, David Luan, and
  Ilya Sutskever.
\newblock Generative pretraining from pixels.
\newblock In {\em International Conference on Machine Learning}, pages
  1691--1703. PMLR, 2020.

\bibitem{cho2014learning}
Kyunghyun Cho, Bart Van~Merri{\"e}nboer, Caglar Gulcehre, Dzmitry Bahdanau,
  Fethi Bougares, Holger Schwenk, and Yoshua Bengio.
\newblock Learning phrase representations using rnn encoder-decoder for
  statistical machine translation.
\newblock {\em arXiv preprint arXiv:1406.1078}, 2014.

\bibitem{choi2016retain}
Edward Choi, Mohammad~Taha Bahadori, Joshua~A Kulas, Andy Schuetz, Walter~F
  Stewart, and Jimeng Sun.
\newblock {RETAIN}: An interpretable predictive model for healthcare using
  reverse time attention mechanism.
\newblock {\em Advances in Neural Information Processing Systems}, pages
  3512--3520, 2016.

\bibitem{choi2016doctor}
Edward Choi, Mohammad~Taha Bahadori, Andy Schuetz, Walter~F Stewart, and Jimeng
  Sun.
\newblock Doctor {AI}: Predicting clinical events via recurrent neural
  networks.
\newblock In {\em Machine Learning for Healthcare Conference}, pages 301--318.
  PMLR, 2016.

\bibitem{choi2019graph}
Edward Choi, Zhen Xu, Yujia Li, Michael~W Dusenberry, Gerardo Flores, Yuan Xue,
  and Andrew~M Dai.
\newblock Graph {C}onvolutional {T}ransformer: Learning the graphical structure
  of electronic health records.
\newblock {\em arXiv preprint arXiv:1906.04716}, 2019.

\bibitem{cueto2019comparative}
Nah{\'u}m Cueto-L{\'o}pez, Maria~Teresa Garc{\'\i}a-Ord{\'a}s, Ver{\'o}nica
  D{\'a}vila-Batista, V{\'\i}ctor Moreno, Nuria Aragon{\'e}s, and Roc{\'\i}o
  Alaiz-Rodr{\'\i}guez.
\newblock A comparative study on feature selection for a risk prediction model
  for colorectal cancer.
\newblock {\em Computer {M}ethods and {P}rograms in {B}iomedicine},
  177:219--229, 2019.

\bibitem{deng2009imagenet}
Jia Deng, Wei Dong, Richard Socher, Li-Jia Li, Kai Li, and Li~Fei-Fei.
\newblock Imagenet: {A} large-scale hierarchical image database.
\newblock In {\em 2009 IEEE {C}onference on {C}omputer {V}ision and {P}attern
  {R}ecognition}, pages 248--255. IEEE, 2009.

\bibitem{devlin2019bert}
Jacob Devlin, Ming-Wei Chang, Kenton Lee, and Kristina Toutanova.
\newblock {BERT}: Pre-training of deep bidirectional transformers for language
  understanding.
\newblock In {\em NAACL-HLT (1)}, 2019.

\bibitem{dosovitskiy2020image}
Alexey Dosovitskiy, Lucas Beyer, Alexander Kolesnikov, Dirk Weissenborn,
  Xiaohua Zhai, Thomas Unterthiner, Mostafa Dehghani, Matthias Minderer, Georg
  Heigold, Sylvain Gelly, et~al.
\newblock An image is worth 16x16 words: Transformers for image recognition at
  scale.
\newblock In {\em International Conference on Learning Representations}, 2020.

\bibitem{ellis2018risk}
Randall~P Ellis, Bruno Martins, and Sherri Rose.
\newblock Risk adjustment for health plan payment.
\newblock In {\em Risk Adjustment, Risk Sharing and Premium Regulation in
  Health Insurance Markets}, pages 55--104. Elsevier, 2018.

\bibitem{fedus2021switch}
William Fedus, Barret Zoph, and Noam Shazeer.
\newblock Switch transformers: {S}caling to trillion parameter models with
  simple and efficient sparsity.
\newblock {\em arXiv preprint arXiv:2101.03961}, 2021.

\bibitem{goldberger2000physiobank}
Ary~L Goldberger, Luis~AN Amaral, Leon Glass, Jeffrey~M Hausdorff, Plamen~Ch
  Ivanov, Roger~G Mark, Joseph~E Mietus, George~B Moody, Chung-Kang Peng, and
  H~Eugene Stanley.
\newblock Physiobank, physiotoolkit, and physionet: components of a new
  research resource for complex physiologic signals.
\newblock {\em circulation}, 101(23):e215--e220, 2000.

\bibitem{Goodfellow-et-al-2016}
Ian Goodfellow, Yoshua Bengio, and Aaron Courville.
\newblock {\em {D}eep {L}earning}.
\newblock MIT Press, 2016.
\newblock \url{http://www.deeplearningbook.org}.

\bibitem{haendel2020many}
Melissa Haendel, Nicole Vasilevsky, Deepak Unni, Cristian Bologa, Nomi Harris,
  Heidi Rehm, Ada Hamosh, Gareth Baynam, Tudor Groza, Julie McMurry, et~al.
\newblock How many rare diseases are there?
\newblock {\em Nature reviews drug discovery}, 19(2):77--78, 2020.

\bibitem{harutyunyan2019multitask}
Hrayr Harutyunyan, Hrant Khachatrian, David~C Kale, Greg Ver~Steeg, and Aram
  Galstyan.
\newblock Multitask learning and benchmarking with clinical time series data.
\newblock {\em Scientific Data}, 6(1):1--18, 2019.

\bibitem{he2016deep}
Kaiming He, Xiangyu Zhang, Shaoqing Ren, and Jian Sun.
\newblock Deep residual learning for image recognition.
\newblock In {\em Proceedings of the {IEEE} {C}onference on {C}omputer {V}ision
  and {P}attern {R}ecognition}, pages 770--778, 2016.

\bibitem{hendrycks2016gaussian}
Dan Hendrycks and Kevin Gimpel.
\newblock Gaussian error linear units ({GELU}s).
\newblock {\em arXiv preprint arXiv:1606.08415}, 2016.

\bibitem{ho2019axial}
Jonathan Ho, Nal Kalchbrenner, Dirk Weissenborn, and Tim Salimans.
\newblock Axial attention in multidimensional transformers.
\newblock {\em arXiv preprint arXiv:1912.12180}, 2019.

\bibitem{hochreiter1997long}
Sepp Hochreiter and J{\"u}rgen Schmidhuber.
\newblock Long short-term memory.
\newblock {\em Neural computation}, 9(8):1735--1780, 1997.

\bibitem{howard2018universal}
Jeremy Howard and Sebastian Ruder.
\newblock Universal language model fine-tuning for text classification.
\newblock {\em arXiv preprint arXiv:1801.06146}, 2018.

\bibitem{thlneedfor}
Unto Häkkinen, Tuukka Holster, Taru Haula, Satu Kapiainen, Petra Kokko, Merja
  Korajoki, Suvi Mäklin, Lien Nguyen, Tuuli Puroharju, and Mikko Peltola.
\newblock Need adjustment for financing health and social services in
  {F}inland. {N}ational {I}nstitute for {H}ealth and {W}elfare ({THL}).

\bibitem{jain2019attention}
Sarthak Jain and Byron~C Wallace.
\newblock Attention is not explanation.
\newblock {\em arXiv preprint arXiv:1902.10186}, 2019.

\bibitem{jiang2023health}
Lavender~Yao Jiang, Xujin~Chris Liu, Nima~Pour Nejatian, Mustafa Nasir-Moin,
  Duo Wang, Anas Abidin, Kevin Eaton, Howard~Antony Riina, Ilya Laufer, Paawan
  Punjabi, et~al.
\newblock Health system-scale language models are all-purpose prediction
  engines.
\newblock {\em Nature}, pages 1--6, 2023.

\bibitem{johnson2020mimic}
A~Johnson, L~Bulgarelli, T~Pollard, S~Horng, LA~Celi, and R~Mark.
\newblock {MIMIC}-{IV} (version 1.0), 2020.

\bibitem{kelly2019key}
Christopher~J Kelly, Alan Karthikesalingam, Mustafa Suleyman, Greg Corrado, and
  Dominic King.
\newblock Key challenges for delivering clinical impact with artificial
  intelligence.
\newblock {\em BMC medicine}, 17:1--9, 2019.

\bibitem{kodialam2021deep}
Rohan Kodialam, Rebecca Boiarsky, Justin Lim, Aditya Sai, Neil Dixit, and David
  Sontag.
\newblock Deep contextual clinical prediction with reverse distillation.
\newblock In {\em Proceedings of the AAAI Conference on Artificial
  Intelligence}, volume~35, pages 249--258, 2021.

\bibitem{krishnan2022self}
Rayan Krishnan, Pranav Rajpurkar, and Eric~J Topol.
\newblock Self-supervised learning in medicine and healthcare.
\newblock {\em Nature Biomedical Engineering}, pages 1--7, 2022.

\bibitem{kumar2020predicting}
Yogesh Kumar, Henri Salo, Tuomo Nieminen, Kristian Vepsalainen, Sangita
  Kulathinal, and Pekka Marttinen.
\newblock Predicting utilization of healthcare services from individual disease
  trajectories using {RNN}s with multi-headed attention.
\newblock In {\em Machine Learning for Health Workshop}, pages 93--111. PMLR,
  2020.

\bibitem{Layton2018EvaluatingTP}
Timothy~J Layton, Randall~P. Ellis, Thomas Mcguire, and Richard~C. van Kleef.
\newblock Evaluating the performance of {H}ealth {P}lan {P}ayment {S}ystems.
\newblock 2018.

\bibitem{lee2021fnet}
James Lee-Thorp, Joshua Ainslie, Ilya Eckstein, and Santiago Ontanon.
\newblock {FN}et: {M}ixing tokens with {F}ourier {T}ransforms.
\newblock {\em arXiv preprint arXiv:2105.03824}, 2021.

\bibitem{thlproc}
Jari Lehtonen, Jukka Lehtovirta, and Päivi Mäkelä-Bengs.
\newblock {THL}-toimenpideluokitus (2013).

\bibitem{li2020behrt}
Yikuan Li, Shishir Rao, Jos{\'e} Roberto~Ayala Solares, Abdelaali Hassaine,
  Rema Ramakrishnan, Dexter Canoy, Yajie Zhu, Kazem Rahimi, and Gholamreza
  Salimi-Khorshidi.
\newblock {BEHRT}: {T}ransformer for electronic health records.
\newblock {\em Scientific Reports}, 10(1):1--12, 2020.

\bibitem{lipton2015learning}
Zachary~C Lipton, David~C Kale, Charles Elkan, and Randall Wetzel.
\newblock Learning to diagnose with {LSTM} recurrent neural networks.
\newblock {\em arXiv preprint arXiv:1511.03677}, 2015.

\bibitem{liu2021pay}
Hanxiao Liu, Zihang Dai, David~R So, and Quoc~V Le.
\newblock Pay attention to {MLP}s.
\newblock {\em Advances in Neural Information Processing Systems}, 2021.

\bibitem{liu2019variance}
Liyuan Liu, Haoming Jiang, Pengcheng He, Weizhu Chen, Xiaodong Liu, Jianfeng
  Gao, and Jiawei Han.
\newblock On the variance of the adaptive learning rate and beyond.
\newblock {\em arXiv preprint arXiv:1908.03265}, 2019.

\bibitem{liu2019roberta}
Yinhan Liu, Myle Ott, Naman Goyal, Jingfei Du, Mandar Joshi, Danqi Chen, Omer
  Levy, Mike Lewis, Luke Zettlemoyer, and Veselin Stoyanov.
\newblock Ro{BERT}a: A robustly optimized {BERT} pretraining approach.
\newblock {\em arXiv preprint arXiv:1907.11692}, 2019.

\bibitem{lundberg2017unified}
Scott~M Lundberg and Su-In Lee.
\newblock A unified approach to interpreting model predictions.
\newblock {\em Advances in {N}eural {I}nformation {P}rocessing {S}ystems}, 30,
  2017.

\bibitem{mcguire2018regulated}
Thomas~G McGuire and Richard~C van Kleef.
\newblock Regulated competition in health insurance markets: Paradigms and
  ongoing issues.
\newblock In {\em Risk Adjustment, Risk Sharing and Premium Regulation in
  Health Insurance Markets}, pages 3--20. Elsevier, 2018.

\bibitem{melas2021you}
Luke Melas-Kyriazi.
\newblock {D}o you even need attention? {A} stack of feed-forward layers does
  surprisingly well on {I}magenet.
\newblock {\em arXiv preprint arXiv:2105.02723}, 2021.

\bibitem{meng2021bidirectional}
Yiwen Meng, William Speier, Michael~K Ong, and Corey~W Arnold.
\newblock Bidirectional representation learning from transformers using
  multimodal electronic health record data to predict depression.
\newblock {\em IEEE Journal of Biomedical and Health Informatics},
  25(8):3121--3129, 2021.

\bibitem{miotto2016deep}
Riccardo Miotto, Li~Li, Brian~A Kidd, and Joel~T Dudley.
\newblock Deep patient: an unsupervised representation to predict the future of
  patients from the electronic health records.
\newblock {\em Scientific reports}, 6(1):1--10, 2016.

\bibitem{mizan2022medical}
Tasquia Mizan and Sharareh Taghipour.
\newblock Medical resource allocation planning by integrating machine learning
  and optimization models.
\newblock {\em Artificial Intelligence in Medicine}, 134:102430, 2022.

\bibitem{openai2023gpt4}
OpenAI.
\newblock {GPT}-4 technical report, 2023.

\bibitem{paszke2017automatic}
Adam Paszke, Sam Gross, Soumith Chintala, Gregory Chanan, Edward Yang, Zachary
  DeVito, Zeming Lin, Alban Desmaison, Luca Antiga, and Adam Lerer.
\newblock Automatic differentiation in {P}ytorch.
\newblock 2017.

\bibitem{peters2018deep}
Matthew~E Peters, Mark Neumann, Mohit Iyyer, Matt Gardner, Christopher Clark,
  Kenton Lee, and Luke Zettlemoyer.
\newblock Deep contextualized word representations.
\newblock {\em arXiv preprint arXiv:1802.05365}, 2018.

\bibitem{radford2019language}
Alec Radford, Jeffrey Wu, Rewon Child, David Luan, Dario Amodei, and Ilya
  Sutskever.
\newblock Language models are unsupervised multitask learners.
\newblock {\em OpenAI blog}, 1(8):9, 2019.

\bibitem{raffel2019exploring}
Colin Raffel, Noam Shazeer, Adam Roberts, Katherine Lee, Sharan Narang, Michael
  Matena, Yanqi Zhou, Wei Li, and Peter~J Liu.
\newblock Exploring the limits of transfer learning with a unified text-to-text
  transformer.
\newblock {\em Journal of Machine Learning Research}, 21:1--67, 2020.

\bibitem{rasmy2021med}
Laila Rasmy, Yang Xiang, Ziqian Xie, Cui Tao, and Degui Zhi.
\newblock Med-{BERT}: {P}retrained contextualized embeddings on large-scale
  structured electronic health records for disease prediction.
\newblock {\em npj Digital Medicine}, 4(1):1--13, 2021.

\bibitem{razavian2015population}
Narges Razavian, Saul Blecker, Ann~Marie Schmidt, Aaron Smith-McLallen, Somesh
  Nigam, and David Sontag.
\newblock Population-level prediction of type 2 diabetes from claims data and
  analysis of risk factors.
\newblock {\em Big Data}, 3(4):277--287, 2015.

\bibitem{rose2016machine}
Sherri Rose.
\newblock A machine learning framework for plan payment risk adjustment.
\newblock {\em Health Services Research}, 51(6):2358--2374, 2016.

\bibitem{russell2021electronic}
Louise~B Russell.
\newblock Electronic health records: the signal and the noise, 2021.

\bibitem{shang2019pre}
Junyuan Shang, Tengfei Ma, Cao Xiao, and Jimeng Sun.
\newblock Pre-training of graph augmented transformers for medication
  recommendation.
\newblock {\em arXiv preprint arXiv:1906.00346}, 2019.

\bibitem{shazeer2020glu}
Noam Shazeer.
\newblock {GLU} variants improve transformer.
\newblock {\em arXiv preprint arXiv:2002.05202}, 2020.

\bibitem{sheen2019metastasis}
Heesoon Sheen, Wook Kim, Byung~Hyun Byun, Chang-Bae Kong, Won~Seok Song,
  Wan~Hyeong Cho, Ilhan Lim, Sang~Moo Lim, and Sang-Keun Woo.
\newblock Metastasis risk prediction model in osteosarcoma using metabolic
  imaging phenotypes: {A} multivariable radiomics model.
\newblock {\em PloS one}, 14(11):e0225242, 2019.

\bibitem{shoeybi2019megatron}
Mohammad Shoeybi, Mostofa Patwary, Raul Puri, Patrick LeGresley, Jared Casper,
  and Bryan Catanzaro.
\newblock Megatron-{LM}: {T}raining multi-billion parameter language models
  using model parallelism.
\newblock {\em arXiv preprint arXiv:1909.08053}, 2019.

\bibitem{shrestha2018mental}
Akritee Shrestha, Savannah Bergquist, Ellen Montz, and Sherri Rose.
\newblock Mental health risk adjustment with clinical categories and machine
  learning.
\newblock {\em Health Services Research}, 53:3189--3206, 2018.

\bibitem{smith2017cyclical}
Leslie~N Smith.
\newblock Cyclical learning rates for training neural networks.
\newblock In {\em 2017 IEEE Winter Conference on Applications of Computer
  Vision (WACV)}, pages 464--472. IEEE, 2017.

\bibitem{tawhid2021machine}
Abdalrahman Tawhid, Tanya Teotia, and Haytham Elmiligi.
\newblock Machine learning for optimizing healthcare resources.
\newblock In {\em Machine Learning, Big Data, and IoT for Medical Informatics},
  pages 215--239. Elsevier, 2021.

\bibitem{tolstikhin2021mlp}
Ilya Tolstikhin, Neil Houlsby, Alexander Kolesnikov, Lucas Beyer, Xiaohua Zhai,
  Thomas Unterthiner, Jessica Yung, Daniel Keysers, Jakob Uszkoreit, Mario
  Lucic, et~al.
\newblock {MLP}-{M}ixer: An all-{MLP} architecture for vision.
\newblock {\em Advances in Neural Information Processing Systems}, 2021.

\bibitem{touvron2021resmlp}
Hugo Touvron, Piotr Bojanowski, Mathilde Caron, Matthieu Cord, Alaaeldin
  El-Nouby, Edouard Grave, Armand Joulin, Gabriel Synnaeve, Jakob Verbeek, and
  Herv{\'e} J{\'e}gou.
\newblock {ResMLP}: Feedforward networks for image classification with
  data-efficient training.
\newblock {\em arXiv preprint arXiv:2105.03404}, 2021.

\bibitem{vaswani2017attention}
Ashish Vaswani, Noam Shazeer, Niki Parmar, Jakob Uszkoreit, Llion Jones,
  Aidan~N Gomez, {\L}ukasz Kaiser, and Illia Polosukhin.
\newblock Attention is all you need.
\newblock In {\em Advances in Neural Information Processing Systems}, pages
  5998--6008, 2017.

\bibitem{wang2018glue}
Alex Wang, Amanpreet Singh, Julian Michael, Felix Hill, Omer Levy, and Samuel~R
  Bowman.
\newblock {GLUE}: A multi-task benchmark and analysis platform for natural
  language understanding.
\newblock {\em arXiv preprint arXiv:1804.07461}, 2018.

\bibitem{wiegreffe2019attention}
Sarah Wiegreffe and Yuval Pinter.
\newblock Attention is not not explanation.
\newblock {\em arXiv preprint arXiv:1908.04626}, 2019.

\bibitem{yan2022clinical}
Bin Yan and Mingtao Pei.
\newblock Clinical-{BERT}: Vision-language pre-training for radiograph
  diagnosis and reports generation.
\newblock In {\em Proceedings of the AAAI Conference on Artificial
  Intelligence}, volume~36, pages 2982--2990, 2022.

\end{thebibliography}

\newpage
\section*{Appendix}

\subsection{SANSformer Components}

\begin{figure}[h]
  \centering
  \includegraphics[width=0.8\linewidth]{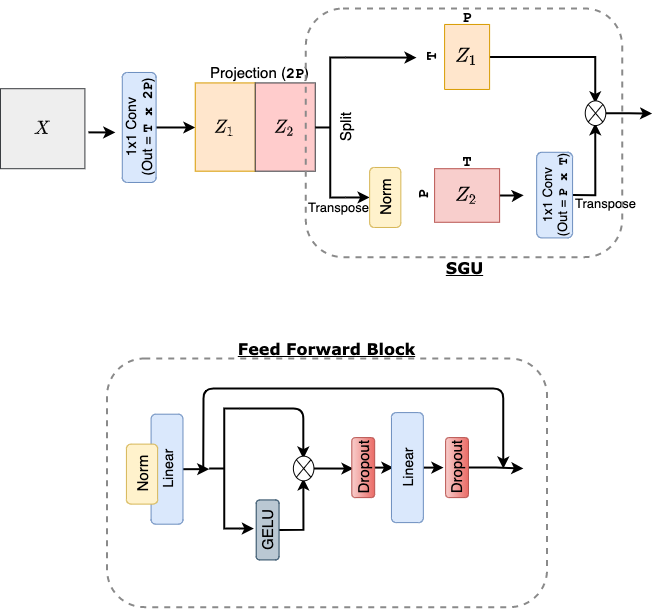}

  \caption[]{Detailed schematic of the Spatial gating unit \emph{Top} and the feedforward blocks \emph{Bottom}. GELU activation is applied throughout the mixer mechanism.  The feedforward (FF) block consists of a series of linear layers with GLU activation and dropout. As explained in \cite{shazeer2020glu}, the GLU operation is applied by splitting the input tensor along the embedding dimension into two chunks, adding GELU activation on one of the chunks and combining them with a multiplicative gate.}
  \label{fig:ff-sgu-fig}
\end{figure}

This section further explains the intricacies of the SANSformer architecture. The Spatial Gating Unit (SGU), shown in Fig. \ref{fig:ff-sgu-fig} \emph{Top}, is crucial for capturing the token interactions across different time steps. Using the GLU variant of activation function for the feedforward block improved the performance of the SANSformer (Fig. \ref{fig:ff-sgu-fig}), however this came with a significant increase in number of model parameters. 

\subsection{Hyperparameter tuning for NN models}
For tuning the hyperparameters of our models, we utilized the Optuna framework \cite{optuna_2019}. This automated process searches through a predefined range or set of values for each parameter to find the combination that results in the best performance on the validation set. We tuned several key parameters, including the learning rate, weight decay, dropout ratio, the number of transformer heads, and $\alpha_{axial}$, which controls the weighting between the axial attention and standard attention in our Axial SANSformer model. The specific ranges are shown in Table \ref{hyper-search-range}.

\begin{table}[H]
  \caption{\textbf{Search ranges for the hyperparameter search.} The best hyperparameters were chosen based on validation loss from 50 trials.}
    \label{hyper-search-range}
  \centering
  \normalsize
  \setlength\tabcolsep{3pt}
  \begin{tabular}{ l l }
    \toprule
    \textbf{Parameters}             & \textbf{Range}      \\
    \midrule
    Learning Rate                   & range [1e-5, 1e-3]  \\

    Weight Decay                    & range [1e-5, 1e-1]  \\

    Dropout Ratio                   & pick [0.1, 0.2, 0.3] \\
    \# of transformer heads         & pick [8, 16]         \\
    $\alpha_{axial}$                & range (0.0, 1.0)         \\

    \bottomrule
  \end{tabular}
  \medskip
\end{table}

Once the best hyperparameters were determined based on the validation set performance, we trained our model using these parameters and evaluated its performance on the test set. This approach ensured that our model was robust and not overfitted to the training data, providing a reliable measure of its potential real-world performance.

\subsection{Ablation Study on Axial SANSformer}
\label{sec: ablation}

We conducted an ablation study for the Axial SANSformer to evaluate the impact of its various components on the model's performance on both MIMIC and PUMMEL datasets. The results are presented in Tables \ref{table: ablation} and \ref{table: pummel-ablation}. Initially, we nullified the effect of axial mixers by setting the parameter $\alpha_{axial}$ to 0, which effectively reverts the model to the Additive SANSformer. Subsequently, we examined the contribution of the $\Delta \tau$ encoding, which encapsulates the temporal difference between two visits. Lastly, we scrutinized the role of positional encoding within the model. Each of these components plays a pivotal role in achieving the optimal performance of the Axial SANSformer. The axial mixers facilitate the processing of visit and code information separately, the $\Delta \tau$ encoding captures essential temporal dynamics between patient visits, and the positional encoding provides crucial sequence order information. Thus for tasks that require the model to capture complex intra-visit interactions, a combination of all these elements is key to maximizing the efficacy of our model in predicting future healthcare utilization. On the other hand, for tasks like Pummel Visits, the additive variant performs better.

\begin{table}[H]
\caption{\textbf{Ablation Study} on MIMIC Mortality task}
\label{table: ablation}
\centering 
\small
\begin{tabular}{l c}
    \toprule
    \multicolumn{1}{c}{\textbf{Model}} & \textbf{AUC \textuparrow} \\
    \midrule
    Axial SANSformer                                                & $0.761 \pm 0.004$\\
    Axial SANSformer ($\alpha_{axial}$ = 0)                         & $0.759 \pm 0.002$\\
    Axial SANSformer (w/o $\Delta \tau$ encoding )                  & $0.760 \pm 0.005$\\
    Axial SANSformer (w/o positional encoding )                     & $0.754 \pm 0.004$\\    
    \bottomrule
\end{tabular}
\end{table}

\begin{table}[H]
\caption{\textbf{Ablation Study} on Pummel Visits task}
\label{table: pummel-ablation}
\centering 
\small
\begin{tabular}{l c}
    \toprule
    \multicolumn{1}{c}{\textbf{Model}} & \textbf{\makecell{Spearman \textuparrow\\ Corr. $y_{\text{count}}$}} \\
    \midrule
    Axial SANSformer                                                & $0.518 \pm 0.003$\\
    Axial SANSformer ($\alpha_{axial}$ = 0)                         & $0.521 \pm 0.003$\\
    Axial SANSformer (w/o $\Delta \tau$ encoding )                  & $0.517 \pm 0.007$\\
    Axial SANSformer (w/o positional encoding )                     & $0.514 \pm 0.002$\\    
    \bottomrule
\end{tabular}
\end{table}

\subsection{Scalability on Pummel}
\label{sec: pummel-scalability}

\begin{figure}[H]
  \centering

      \includegraphics[scale=0.65]{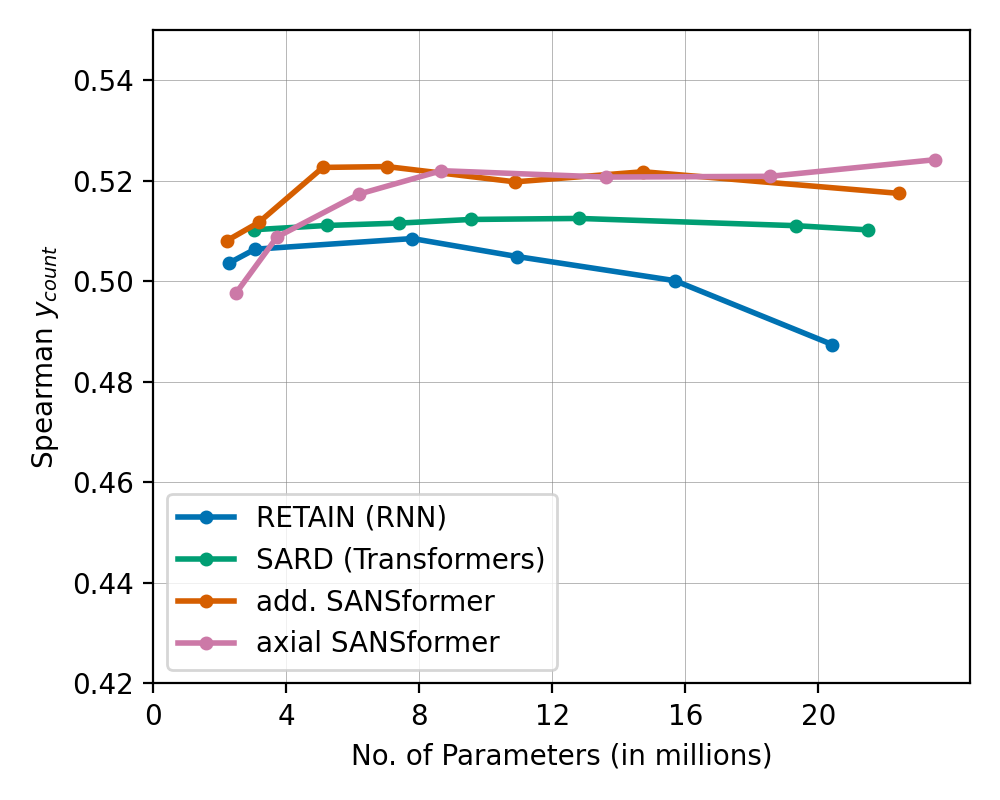}
      \caption[]{\textbf{Parameter Scalability on Pummel Visits.} This figure plots model performance on Task 1 against the number of trainable parameters, illustrating the scalability of each model. Both the additive and axial SANSformer models maintain strong performance even as the number of parameters is increased. }
      \label{fig:pummel-param}
\end{figure}

In line with our exploration of SANSformers' scalability on the MIMIC dataset, we extended the scalability study to the Pummel visits task. As in our main experiments, we consistently increase the number of parameters in the models, aiming to assess the performance enhancements, if any, associated with larger model sizes. The scalability performance curve is shown in Fig. \ref{fig:pummel-param}. Given that the training data size for this task is three times that of MIMIC, the resulting scalability curves on the Pummel dataset aren't as distinctly resolved as those observed on the MIMIC dataset. However, SANSformers' demonstrate a similar resistance to overfitting even as the model size grows, underscoring the model's adaptability and stability across diverse datasets.

\subsection{A closer look at the Multiple Sclerosis subgroup}
\label{ms-error-analysis}

In our analysis, we observed that the performances, as reflected by the Spearman correlation and MAE metrics, of the Lasso, SARD, and Axial SANSformer models were closely matched when evaluated on the Multiple Sclerosis (MS) subgroup (Table \ref{divergent-results} \textit{Below}).  A potential cause for these close empirical results is the relatively small test set size for the MS subgroup, which comprises only 27 samples. Such a limited sample size might introduce inherent variability, influencing the observed differences in model performance.

To assess the statistical significance of the MAE scores among the models, we utilized bootstrapping—we repeatedly resample the test set with replacement. This provided a distribution of MAE scores for each model, capturing potential variability from the limited sample size. Subsequent \textit{Wilcoxon signed-rank tests} between Lasso and Axial SANSformer, as well as between SARD and Axial SANSformer, yielded p-values well below 0.005, indicating statistically significant differences. With this established, we delved into further visual analyses to deepen our understanding.

\begin{figure}[h]
  \centering
  \includegraphics[width=\linewidth]{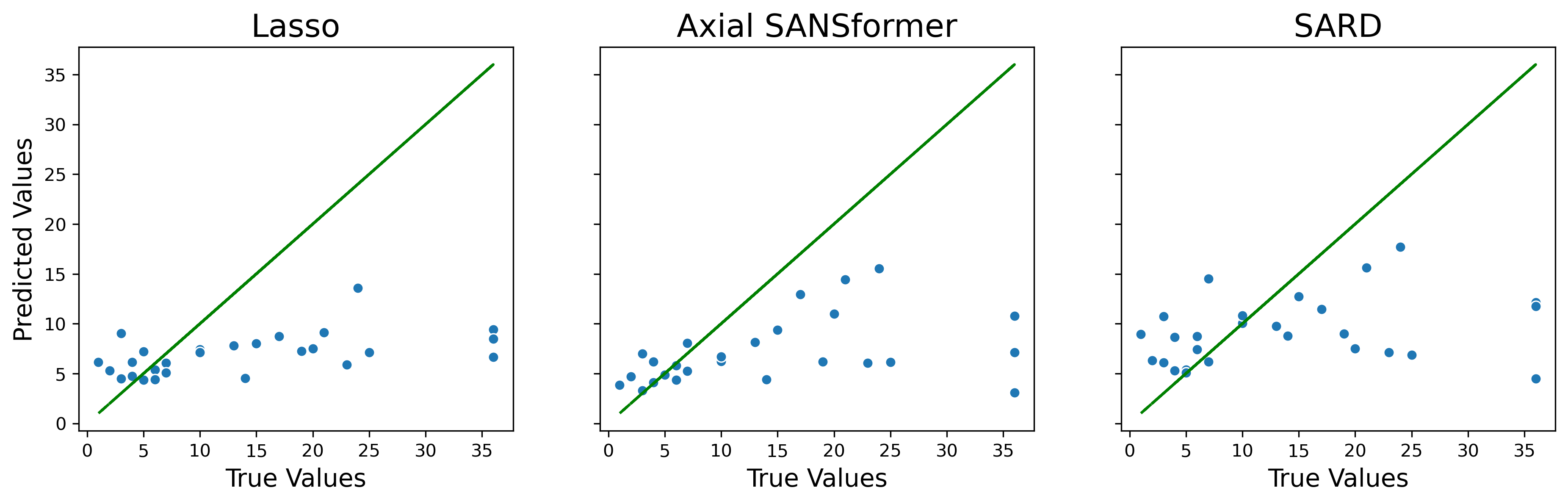}
    \caption[]{\textbf{Scatter Plot Comparing Model Predictions}. The scatter plots display the predictions of the Lasso, SARD, and Axial SANSformers against the true values. While a majority of predictions align closely with the true counts, indicating accurate performance, all models exhibit a discernible underestimation trend for higher count values. }
  \label{fig:scatter-ms}
\end{figure}

To visually compare the prediction patterns of the Lasso, SARD, and Axial SANSformer models, we plotted scatter diagrams, juxtaposing the predictions of each model against the others on the test set as shown in Fig. \ref{fig:scatter-ms}. These diagrams underscore that, for a significant portion of the data, predictions are well-ordered with the actual counts (as also confirmed by the Spearman Rank Correlation from Table \ref{divergent-results} ). However, a recurring challenge observed across all models is the underestimation of higher count values, highlighting areas for potential model improvement.

\begin{figure}[h]
  \centering
  \includegraphics[width=\linewidth]{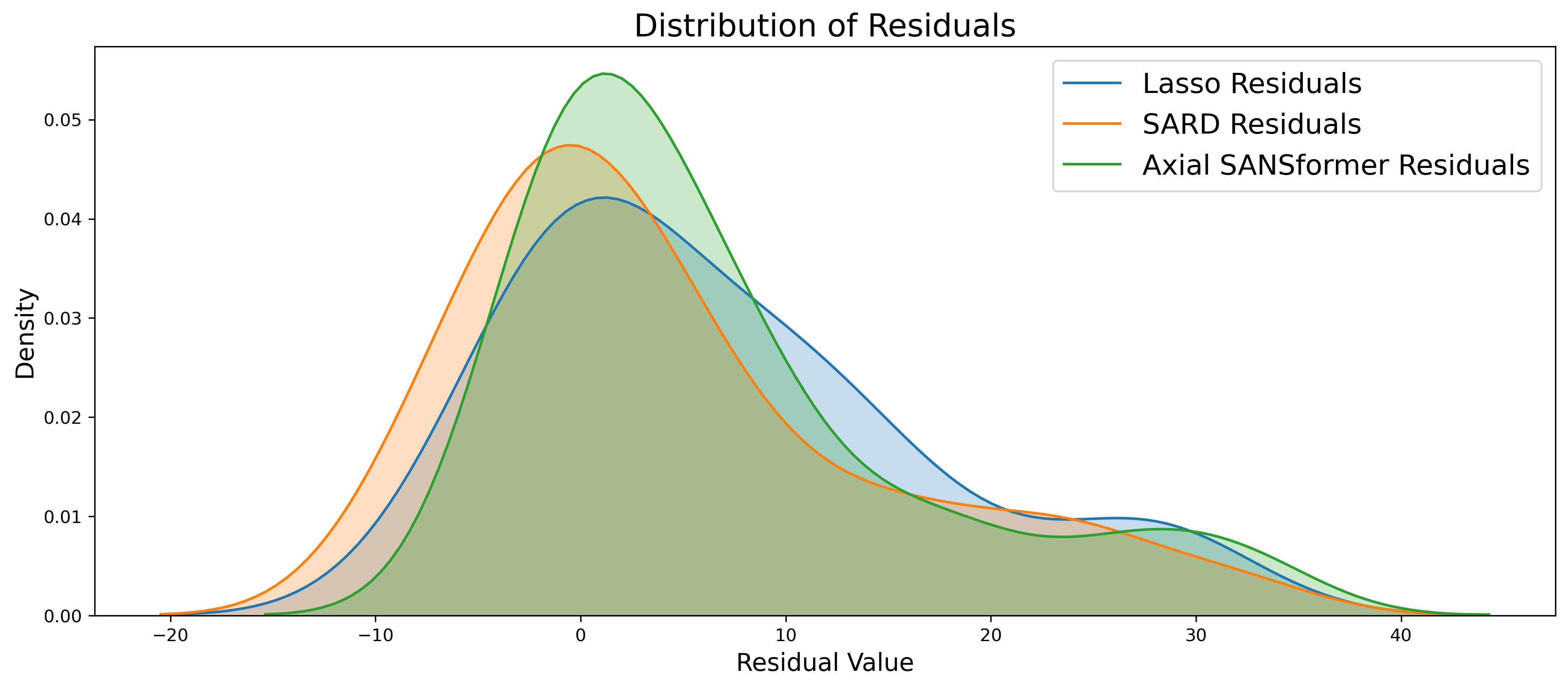}
    \caption[]{\textbf{Kernel Density Estimation (KDE) of Model Residuals}. The KDE plots showcase the distribution of residuals for Lasso, SARD, and Axial SANSformers. All models show a pronounced peak near zero, indicating frequent accurate predictions. Among the three, Axial SANSformers present the highest peak, suggesting a slightly more consistent prediction accuracy, with SARD and Lasso following in that order.}
  \label{fig:kde-residual-ms}
\end{figure}

To gain deeper insights into the predictive behaviors of our models, we conducted a residual analysis. Investigating the distribution of residuals—the differences between the observed and predicted values—sheds light on the patterns of errors and potential areas of improvement. The Kernel Density Estimation (KDE) plots of these residuals, presented in Fig. \ref{fig:kde-residual-ms}, reveal that while all models generally predict close to the true values, they occasionally face challenges, especially with estimating higher count values. This observation is particularly underscored by the long tails evident in the plots.

\begin{figure}[h]
  \centering
  \includegraphics[width=\linewidth]{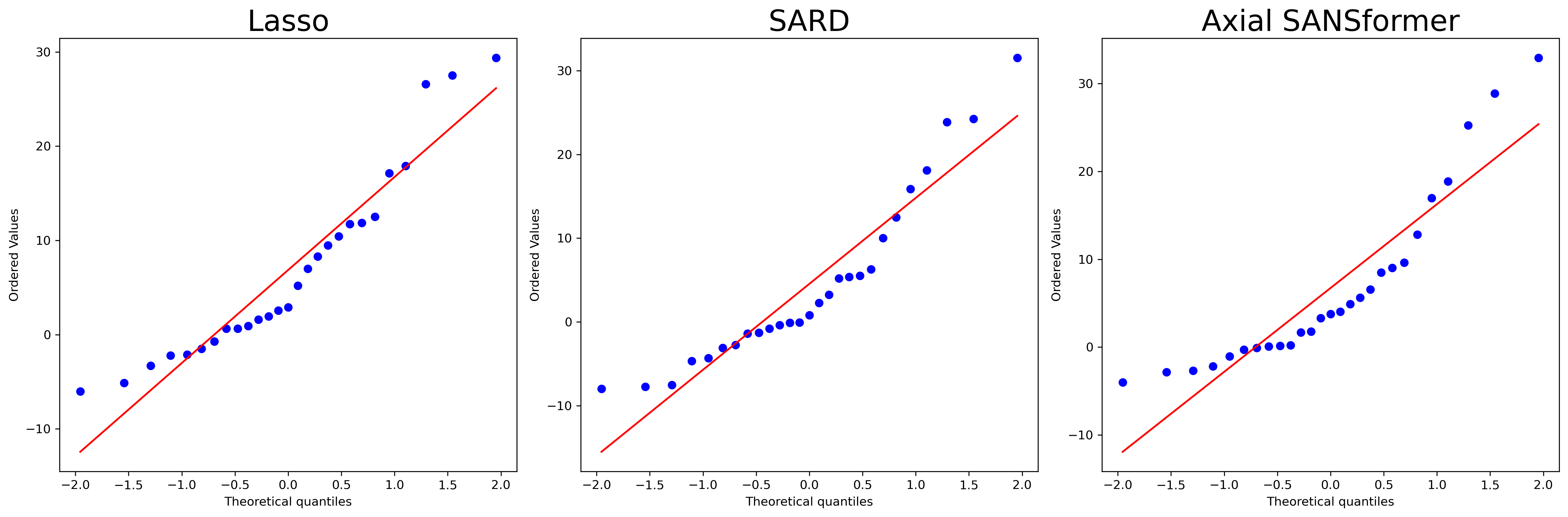}
    \caption[]{\textbf{Q-Q Plot of Model Residuals}. The Q-Q plots provide a visual assessment of how the residuals from the Lasso, SARD, and Axial SANSformers compare to a standard normal distribution. While the central portions of the plots align closely with the diagonal, indicating approximate normality, deviations in the tails, especially the upper quantiles, suggest the presence of heavier tails than expected under normality. This is consistent with the models' occasional struggle with high count values, as observed in other visual diagnostics.}

  \label{fig:qq-residual-ms}
\end{figure}

To further assess the normality of our model residuals, we employed the Q-Q (Quantile-Quantile) plots which juxtaposes the quantiles of the residuals against the quantiles of a standard normal distribution. As depicted in Fig. \ref{fig:qq-residual-ms}, the residuals for the central portions of the data are fairly normally distributed, adhering closely to the diagonal. However, deviations in the tails, particularly the upper quantiles, point to the presence of larger errors than expected under a perfect normal distribution. This observation echoes our earlier findings from both scatter and KDE plots, where the models exhibited challenges in predicting extreme values.

\subsubsection*{Model Recommendation}
In our detailed analysis of the Multiple Sclerosis subgroup, several insights emerge. As depicted in Table \ref{divergent-results}, the SARD model achieves the lowest MAE, aligning with our study's primary objective of predicting healthcare utilization. However, we should note that the MAE metric is prone to be influenced by outliers, hence the Spearman rank correlation is a better metric in such situations (Table \ref{divergent-results}). Concurrently, the Axial SANSformer showcases favorable residuals, evident from the KDE and Q-Q plots. Despite these performances, the modest training sample size (123 samples) would compel us to pick the Lasso model, which combines competitive performance with the advantages of interpretability and feature selection on this subgroup. Nonetheless, the enhanced performance of complex neural models, such as SARD and SANSformers, underscores the significant impact of our Generative Summary Pretraining (GSP) strategy.

\newpage
\onecolumn
\subsection{Disease-specific Task 2 metrics for divergent subgroup}

In the main section, we presented an aggregated evaluation of our model's performance across six significant disease categories in Finland, namely: cancer (ICD-10 codes starting with C and some D), endocrine and metabolic diseases (E), diseases related to nervous systems (G), diseases of the circulatory system (I), diseases of the respiratory system (J), and diseases of the digestive system (K). This aggregation was necessary due to space constraints in the main text. However, it is important to consider that each disease category has unique characteristics and may influence the predictive performance of our models differently. Therefore, in this supplementary section, we provide a more granular view of our model's performance by breaking down the results for each individual disease category.

First, we evaluate our model's performance on predicting the healthcare utilization for each of the six disease categories without pretraining. This would be analogous to the results presented in Table \ref{ri-results-task2} of the main text, but with the results stratified by disease category. Next, we delve into our model's transfer learning capabilities by evaluating its performance on two specific patient subgroups - those diagnosed with Bipolar disorder and Multiple Sclerosis. These results are shown in Tables \ref{table: t2d-split-task2}, \ref{table: bp-split-task2} and \ref{table: ms-split-task2}.

\begin{table*}[ht]
  \caption{\textbf{Diseases specific results for Pummel Diagnoses task.} The model was trained on the Type 2 diabetes subgroup ($N$=41,761). We report the mean $\pm$ standard deviation of the runs from 5 random restarts. `Add.' indicates additive visit summarizer. The diseases are represented by the first character of their respective ICD-10 codes. The cummulative results, with the mean across each disease category, is presented in the Table \ref{ri-results-task2}.}
  \label{table: t2d-split-task2}
  \centering
  \small
\begin{tabular}{@{}lcccccc@{}}
\toprule
\multicolumn{1}{c}{\textbf{Model}} & \multicolumn{6}{c}{\textbf{Spearman Corr. $y_{\text{diag}}$}} \\ \cmidrule(l){2-7}
& \textbf{C, D} & \textbf{E} & \textbf{G} & \textbf{I} & \textbf{J} & \textbf{K} \\ \midrule

RETAIN \cite{choi2016retain}    & 0.338 $\pm$ 0.005 & 0.183 $\pm$ 0.006 & 0.201 $\pm$ 0.020 & 0.318 $\pm$ 0.008 & 0.254 $\pm$ 0.007 & 0.197 $\pm$ 0.020 \\
BEHRT \cite{li2020behrt}        & 0.352 $\pm$ 0.009 & 0.153 $\pm$ 0.002 & 0.175 $\pm$ 0.005 & 0.305 $\pm$ 0.009 & 0.220 $\pm$ 0.004 & 0.146 $\pm$ 0.005 \\
BRLTM \cite{meng2021bidirectional}    & 0.393 $\pm$ 0.004 & 0.162 $\pm$ 0.002 & 0.211 $\pm$ 0.011 & 0.335 $\pm$ 0.006 & 0.253 $\pm$ 0.012 & 0.202 $\pm$ 0.008 \\
SARD \cite{kodialam2021deep}    & 0.320 $\pm$ 0.044 & 0.174 $\pm$ 0.023 & 0.193 $\pm$ 0.027 & 0.306 $\pm$ 0.016 & 0.222 $\pm$ 0.029 & 0.183 $\pm$ 0.025 \\
Add. SANSformer (ours)          & 0.376 $\pm$ 0.004 & 0.197 $\pm$ 0.004 & 0.196 $\pm$ 0.006 & 0.341 $\pm$ 0.006 & 0.257 $\pm$ 0.006 & 0.192 $\pm$ 0.008 \\
Axial SANSformer (ours)         & 0.380 $\pm$ 0.005 & 0.199 $\pm$ 0.003 & 0.197 $\pm$ 0.008 & 0.341 $\pm$ 0.004 & 0.260 $\pm$ 0.012 & 0.193 $\pm$ 0.007 \\ \bottomrule
\end{tabular}
\end{table*}

\begin{table*}[ht]
  \caption{\textbf{Diseases specific results for Pummel Diagnoses task on bipolar disorder subgroup} We report the mean $\pm$ standard deviation of the runs from 5 random restarts. The models were initialized with the weights from Generative Summary Pretraining (GSP). `Add.' indicates additive visit summarizer. The diseases are represented by the first character of their respective ICD-10 codes. The cummulative results, with the mean across each disease category, is presented in the Table \ref{divergent-results} \emph{Top}}
  \label{table: bp-split-task2}
  \centering
  \small
\begin{tabular}{@{}lcccccc@{}}
\toprule
\multicolumn{1}{c}{\textbf{Model}} & \multicolumn{6}{c}{\textbf{Spearman Corr. $y_{\text{diag}}$}} \\ \cmidrule(l){2-7}
& \textbf{C, D} & \textbf{E} & \textbf{G} & \textbf{I} & \textbf{J} & \textbf{K} \\ \midrule

RETAIN + GSP    & 0.282 $\pm$ 0.025 & 0.388 $\pm$ 0.010 & 0.294 $\pm$ 0.026 & 0.392 $\pm$ 0.018 & 0.317 $\pm$ 0.011 & 0.257 $\pm$ 0.039 \\
BEHRT + MLM    & 0.278 $\pm$ 0.029 & 0.254 $\pm$ 0.030 & 0.250 $\pm$ 0.026 & 0.355 $\pm$ 0.036 & 0.255 $\pm$ 0.053 & 0.280 $\pm$ 0.042 \\
BRLTM + MLM    & 0.286 $\pm$ 0.066 & 0.289 $\pm$ 0.020 & 0.247 $\pm$ 0.032 & 0.392 $\pm$ 0.039 & 0.266 $\pm$ 0.017 & 0.307 $\pm$ 0.022 \\
SARD + GSP    & 0.312 $\pm$ 0.009 & 0.301 $\pm$ 0.025 & 0.283 $\pm$ 0.006 & 0.416 $\pm$ 0.013 & 0.277 $\pm$ 0.026 & 0.243 $\pm$ 0.019 \\
Add. SANSformer + GSP          & 0.184 $\pm$ 0.011 & 0.158 $\pm$ 0.025 & 0.184 $\pm$ 0.020 & 0.271 $\pm$ 0.018 & 0.120 $\pm$ 0.006 & 0.170 $\pm$ 0.019 \\
Axial SANSformer + GSP         & 0.343 $\pm$ 0.016 & 0.441 $\pm$ 0.012 & 0.302 $\pm$ 0.010 & 0.479 $\pm$ 0.019 & 0.335 $\pm$ 0.010 & 0.330 $\pm$ 0.015 \\ \bottomrule
\end{tabular}
\end{table*}

\begin{table*}[ht]
    \caption{\textbf{Diseases specific results for Pummel Diagnoses task on Multiple Sclerosis subgroup} We report the mean $\pm$ standard deviation of the runs from 5 random restarts. The models were initialized with the weights from Generative Summary Pretraining (GSP). `Add.' indicates additive visit summarizer. The diseases are represented by the first character of their respective ICD-10 codes. The cummulative results, with the mean across each disease category, is presented in the Table \ref{divergent-results} \emph{Bottom}}
  \label{table: ms-split-task2}
  \centering
  \small
\begin{tabular}{@{}lcccccc@{}}
\toprule
\multicolumn{1}{c}{\textbf{Model}} & \multicolumn{6}{c}{\textbf{Spearman Corr. $y_{\text{diag}}$}} \\ \cmidrule(l){2-7}
& \textbf{C, D} & \textbf{E} & \textbf{G} & \textbf{I} & \textbf{J} & \textbf{K} \\ \midrule

RETAIN + GSP            & 0.423 $\pm$ 0.028 & 0.226 $\pm$ 0.026 & 0.010 $\pm$ 0.033 & 0.440 $\pm$ 0.006 & 0.087 $\pm$ 0.017 & 0.196 $\pm$ 0.024 \\

BEHRT + MLM              & 0.207 $\pm$ 0.096 & 0.069 $\pm$ 0.033 & 0.300 $\pm$ 0.045 & 0.028 $\pm$ 0.056 & 0.194 $\pm$ 0.133 & 0.015 $\pm$ 0.084 \\
BRLTM + MLM              & 0.301 $\pm$ 0.042 & 0.030 $\pm$ 0.033 & 0.300 $\pm$ 0.045 & 0.028 $\pm$ 0.056 & 0.263 $\pm$ 0.070 & 0.119 $\pm$ 0.059 \\
SARD + GSP              & 0.206 $\pm$ 0.038 & 0.142 $\pm$ 0.038 & 0.259 $\pm$ 0.016 & 0.401 $\pm$ 0.064 & 0.152 $\pm$ 0.020 & 0.277 $\pm$ 0.026 \\

Add. SANSformer + GSP   & 0.207 $\pm$ 0.042 & 0.206 $\pm$ 0.100 & 0.002 $\pm$ 0.051 & 0.607 $\pm$ 0.032 & 0.190 $\pm$ 0.138 & 0.441 $\pm$ 0.067 \\

Axial SANSformer + GSP  & 0.285 $\pm$ 0.020 & 0.318 $\pm$ 0.049 & 0.032 $\pm$ 0.027 & 0.615 $\pm$ 0.026 & 0.180 $\pm$ 0.034 & 0.364 $\pm$ 0.060 \\ \bottomrule
\end{tabular}
\end{table*}

\twocolumn
\end{document}